%% file: main.tex
\documentclass{article} 
\usepackage{iclr2024_conference,times}
\vspace{0pt}
\usepackage{parskip}
\input{math_commands.tex}

\usepackage{hyperref}
\usepackage{url}
\usepackage{booktabs}
\usepackage{multirow}
\usepackage{graphicx}
\usepackage{float}
\usepackage{titlesec}
\usepackage[ruled,vlined]{algorithm2e}
\usepackage{caption}
\usepackage{colortbl}
\usepackage{xcolor}
\usepackage{subcaption}
\usepackage{wrapfig}
\usepackage{float}
\usepackage{helvet}
\input{macros.tex}
\definecolor{grayrowcolor}{RGB}{220,220,220} 

\def\shownotes{1}  
\ifnum\shownotes=1
\newcommand{\authnote}[2]{[#1: #2]}
\else
\newcommand{\authnote}[2]{}
\fi

\SetAlgoNlRelativeSize{0}

\definecolor{darkpink}{RGB}{255, 105, 180} 

\title{Llamas Know What GPTs Don't Show: \\Surrogate Models for Confidence Estimation}

\newcommand{\authorsep}{\hspace{2ex}}

\author{Vaishnavi Shrivastava,
\authorsep 
Percy Liang, 
\authorsep 
Ananya Kumar\\\\
Stanford University\\\texttt{{\{vshrivas, pliang, ananya\}@cs.stanford.edu}}
}

\iclrfinalcopy 
\begin{document}

\maketitle

\input{sections/abstract}
\input{sections/introduction}

\input{sections/setup}
\input{sections/methods}

\input{sections/results/linguistic_confidence_results}
\input{sections/results/surrogate_mixture_confidence_results}

\input{sections/results/analysis}
\input{sections/related_work}
\input{sections/conclusion}
\newpage
\bibliography{iclr2024_conference}
\bibliographystyle{iclr2024_conference}
\newpage
\input{sections/appendix}
\end{document}

%% file: math_commands.tex
\usepackage{amsmath,amsfonts,bm}

\def\eqref#1{equation~\ref{#1}}

\def\1{\bm{1}}

\DeclareMathAlphabet{\mathsfit}{\encodingdefault}{\sfdefault}{m}{sl}
\SetMathAlphabet{\mathsfit}{bold}{\encodingdefault}{\sfdefault}{bx}{n}

\newcommand{\E}{\mathbb{E}}



%% file: macros.tex
\usepackage{amsthm}
\usepackage[english]{babel}
\usepackage[T1]{fontenc}
\usepackage{amsmath}
\usepackage{amssymb}
\usepackage{amsfonts}
\usepackage{multirow}
\usepackage{mathtools}
\usepackage{breqn}
\usepackage{bbm}
\usepackage{dsfont}
\usepackage{graphicx}
\usepackage{natbib}
\usepackage{hyperref}
\usepackage{url}
\usepackage{cases}
\usepackage[colorinlistoftodos]{todonotes}
\usepackage{booktabs}
\usepackage{array}
\usepackage{paralist}
\usepackage{wrapfig}

\newcommand{\correct}{R}
\newcommand{\conf}{C}

\newcommand{\auc}{\mbox{AUC}}

\renewcommand{\hat}{\widehat}

\newtheorem*{remark*}{Remark}

\newtheorem*{observation*}{Observation}

\numberwithin{equation}{section}

%% file: sections/abstract.tex
\begin{abstract}
To maintain user trust, large language models (LLMs) should signal low confidence on examples where they are incorrect, instead of misleading the user.
The standard approach of estimating confidence is to use the softmax probabilities of these models, but as of November 2023, state-of-the-art LLMs such as GPT-4 and Claude-v1.3 do not provide access to these probabilities.
We first study eliciting confidence linguistically --- asking an LLM for its confidence in its answer --- which performs reasonably (80.5\% AUC on GPT-4 averaged across 12 question-answering datasets --- 7\% above a random baseline) but leaves room for improvement.
We then explore using a \emph{surrogate} confidence model --- using a model where we do have probabilities to evaluate the \emph{original} model's confidence in a given question.
Surprisingly, even though these probabilities come from a different and often weaker model, this method leads to higher AUC than linguistic confidences on
9 out of 12 datasets.
Our best method composing linguistic confidences and surrogate model probabilities gives state-of-the-art confidence estimates
on all 12 datasets (84.6\% average AUC on GPT-4).
\end{abstract}

%% file: sections/introduction.tex
\section{Introduction}

As large language models (LLMs) are increasingly deployed, it is important that they signal low confidence on examples where they are likely to make mistakes.
This problem is called selective classification (or classification with a reject option) and is widely studied in machine learning~\citep{cordella1995method,geifman2017selective,feng2019selective,jones2021selective}, learning theory~\citep{elyaniv2010foundations,bartlett2008classification}, and natural language processing~\citep{kamath2020squads,liang2022helm,xiong2023can}.
Traditional approaches leverage the model's softmax probabilities~\citep{hendrycks2017baseline,jones2021selective,liang2022helm} or the model's representations~\citep{lee2018unified}.
This paper’s goal is to produce \emph{good confidence estimates for state-of-the-art LLMs}, which \emph{do not provide model probabilities or representations} (such as GPT-4 and Claude-v1.3).

\begin{figure*}[t]
    \centering
    \includegraphics[width=\textwidth]
{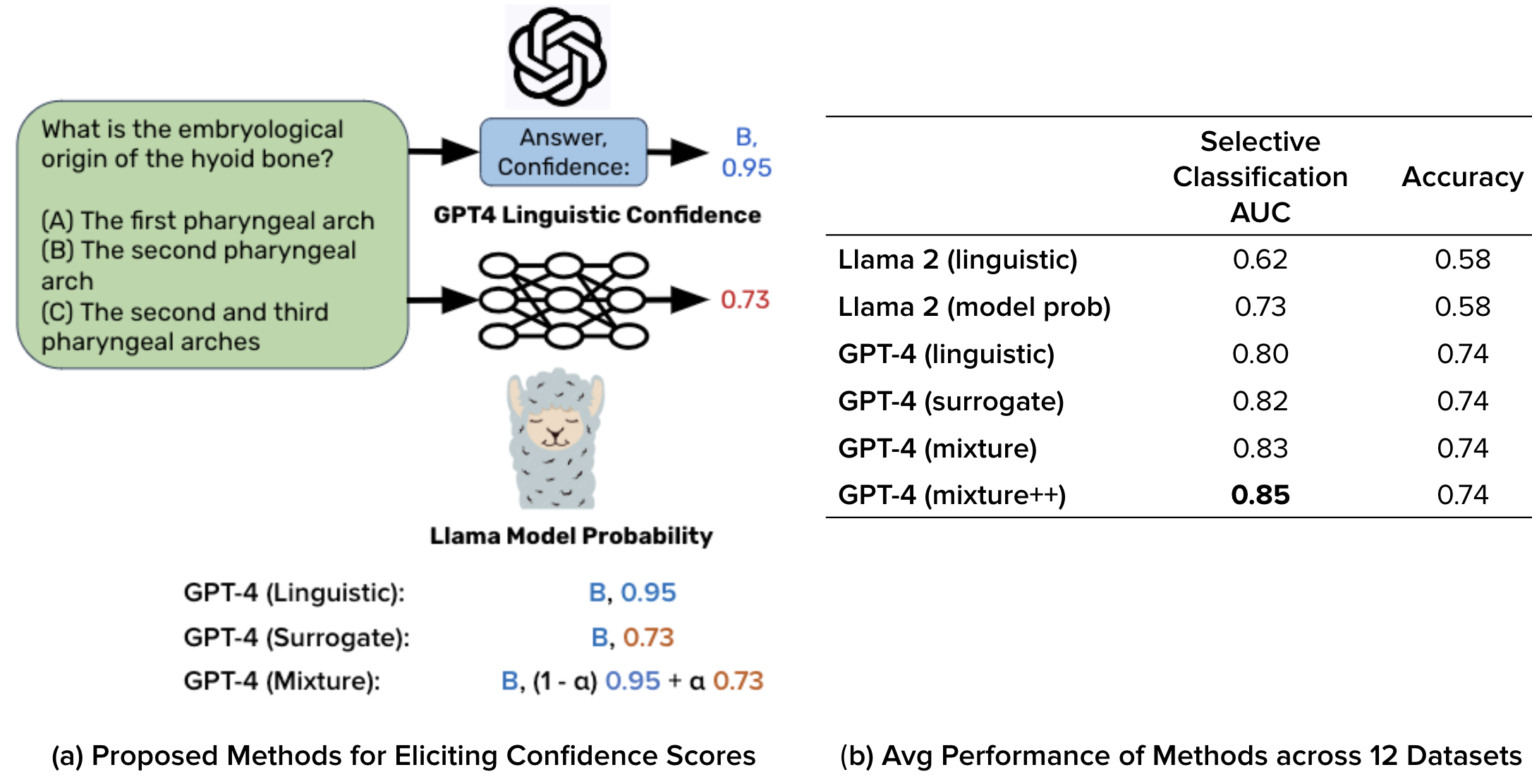}
    \caption{
      Our goal is to provide good confidence estimates for state-of-the-art LLMs like GPT-4 and Claude-v1.3 which currently do not give access to their internal probabilities.
      One natural approach (GPT-4 Linguistic) is to prompt the model asking for its confidence.
      Interestingly, we find that taking the answer from GPT-4, but the internal probability from a different surrogate model (e.g., an open model such as Llama 2) gives even better results (0.82 AUC).
      Mixing GPT-4's linguistic confidences with the surrogate model probabilities gives further gains (0.83 AUC).
      Our AUC numbers are better than concurrent work~\citep{xiong2023can}, but combining these approaches leads to the best results (Mixture++; 0.85 AUC).
      Our findings also hold for Claude-v1.3 and GPT-3.5 (Section~\ref{sec:surrogate_confidence_models} and~\ref{sec:mixtures_confidence_models}).
    }
    \label{fig:intro-fig}
\end{figure*}

We first examine a natural idea of eliciting linguistic confidence scores~\citep{tian2023just,lin2022teaching,xiong2023can} --- prompting the LLM
to assess its confidence in its answer (Figure~\ref{fig:intro-fig}, GPT-4 Linguistic).
We find that linguistic confidences work reasonably well for state-of-the-art models, and much better than a random guessing baseline, but still leave room for improvement (Section~\ref{sec:ling_confs}). Averaged across the datasets, GPT-4 achieves 
a selective classification AUC of 80.5\%, which is 7\% above a random guessing baseline.
Our results hold across 12 standard datasets (8 MMLU datasets, TruthfulQA, CommonsenseQA, OpenbookQA, and MedQA), 5 models (GPT-4, Claude-v1.3, GPT-3.5, Llama 2, and text-davinci-003), and 24 different prompt formats (e.g., chain-of-thought, different instructions, fake few-shot prompts).
However, linguistic confidences perform much worse than using model probabilities when these probabilities are available (for less accurate models).
For example, on Llama 2 linguistic confidences achieve an average AUC 10.7\% lower than model probabilities, suggesting scope for further refinement in these confidence assessments.

Consequently, we propose a surrogate model approach of taking the answer from GPT-4 or Claude-v1.3, but the \emph{confidence from a different model} such as Llama 2 (Figure~\ref{fig:intro-fig}, Surrogate), where softmax probabilities are available, as a confidence estimate
for the original model's answer (Section ~\ref{sec:surrogate_confidence_models}).
Surrogate confidence modeling improves the average selective classification AUC for GPT-4 to 82.1\%. 
Even using a weaker or
much smaller surrogate model like text-davinci-003 or Llama 2-13B 
leads to comparable or better AUCs for stronger models such as GPT-4, Claude-v1.3, and GPT-3.5. 
Intriguingly, confidence scores can transfer between models, even if the model generating the confidence score is different (or much worse).
In Section ~\ref{sec:surrogate_confidence_models}, we provide some analysis and intuitions for this behavior.

We find that linguistic confidence scores  
and surrogate model probabilities are complementary:
combining these scores 
leads to further gains (Figure~\ref{fig:intro-fig}, Mixture).
For example, this mixture method
increases the selective classification AUC of GPT-4 to 83.4\%.
The mixture method also outperforms concurrent work~\citep{xiong2023can} on self-consistency 
(AUC: 82.8\%),
which is more expensive (involves sampling GPT-4 five times per input) and involves post-processing.
Combining our method with self-consistency-based confidence scores leads to the \emph{best results: average AUC
84.6\%}.

Our analysis suggests that linguistic confidence scores
are limited
because they are very coarse-grained --- for example, GPT-4 outputs the exact same confidence (0.9) on 50\% of examples, which constrains its ability to separate correct and incorrect answers.
Surrogate model probabilities work well even on a different model, because the examples that are challenging for one model transfer over to a different model.
Finally, mixing in just a small fraction of surrogate model probabilities allows answers which previously had the same linguistic confidence to be separable through different composite confidence scores, boosting the overall performance with minimal interventions.

%% file: sections/setup.tex
\section{Setup}
Our goal is selective classification: outputting confidence scores that are higher on inputs where the model is correct, than inputs where the model is incorrect~\citep{elyaniv2010foundations,geifman2017selective}.
We focus on state-of-the-art language models such as GPT-4 and Claude-v1.3, which currently do not expose probabilities computed in their softmax output layer.

\textbf{Task.}
Given a text input $x$, a model outputs a (possibly stochastic) answer $y(x)$.
Let $\correct(x, y) = 1$ if an answer $y$ is correct for input $x$, and $0$ otherwise.
Our goal is to output a \emph{confidence score} $\conf(x) \in [0, 1]$.
Good confidence scores are essential in real world machine learning systems: for inputs when $\conf(x)$ is lower, we can defer to a human expert or alert the user, instead of misleading the user with an incorrect answer.

\textbf{Metrics.} A popular metric for selective classification is the \emph{AUC} (area under the coverage-accuracy curve)~\citep{elyaniv2010foundations,liang2022helm}, which examines how accurate the model is if allowed to abstain (say "I don't know") on some examples.
Let $A(c)$ be the selective accuracy at coverage $c$: the accuracy if the model only makes a prediction on the $c$ proportion of  data with highest confidence scores.
To enable tie-breaking to make different predictions for examples with the same confidence score,
we add a small amount of Gaussian noise to each confidence score $\mathcal{N}(0, \epsilon), \epsilon \to 0$.
The AUC is the average selective accuracy $A(c)$ over all $c$:
\begin{equation}
    \auc(\conf, y) = 
    \lim_{\epsilon \to 0} \int_0^1 \E\left[A(c)  \right] dc
\end{equation}
A random baseline (outputting uniform random probabilities for each input) achieves $\auc(\conf, y) = \mbox{accuracy}$, so a model with good confidence scores should achieve a higher AUC than accuracy. Note that adding the noise $\mathcal{N}(0, \epsilon)$ is critical because linguistic confidences for different examples are often identical --- without the noise we would substantially underestimate the AUC of the models (see Appendix ~\ref{sec:auc_auroc_definitions} for more details).

We also examine the \emph{AUROC}, a standard metric~\citep{hendrycks2017baseline,xiong2023can} used to examine how well confidence scores can distinguish between correct and incorrect examples.
We label an example `Positive' if the model gets it correct and `Negative' otherwise, and plot the true positive rate against the false positive rate at different classification thresholds --- the AUROC is the area under this curve (See Appendix ~\ref{sec:auc_auroc_definitions} for more details).
Outputting random confidence scores gets an AUROC of 0.5, so a model with good confidence scores should achieve AUROC above 0.5.

We also report \emph{ECE (expected calibration error)} numbers in Appendix~\ref{sec:ece_results}. ECE examines if a model's confidence aligns with its accuracy, but does not indicate the model's ability to distinguish between correct and incorrect examples, so we focus on the AUC and AUROC metrics.\footnote{Intuitively, calibration requires that if we output a $0.6$ confidence on $100$ examples, then we should get $0.6 \cdot 100 = 60$ of them correct. For a classifier with accuracy $A$, one (degenerate) way to have perfect calibration (best possible ECE) is to output confidence $\conf(x) = A$ for every example $x$.}

\textbf{Datasets.}
We study model performance and confidence on twelve standard question answering datasets: \textit{TruthfulQA} (TQA) ~\citep{lin2021truthful}, \textit{CommonsenseQA} (CSQA) ~\citep{talmor2019commonsenseqa}, \textit{OpenbookQA} (OBQA) ~\citep{mihaylov2018openbook}, \textit{MedQA} ~\citep{jin2021medqa}, and 8 \textit{MMLU}~\citep{hendrycks2021measuring} datasets - professional law (Law), business ethics (Ethics), conceptual physics (Physics), econometrics (Econ), abstract algebra (Algebra), college chemistry (Chem), computer security (Security), and US Foreign Policy (Policy). These datasets span several diverse categories including math reasoning, scientific knowledge, computer science, social science, and commonsense reasoning. We sample 250 questions from the test split of each dataset to report results on (if the test set is smaller, we use the full test set). See Appendix \ref{sec:dataset_details} for more details.

\textbf{Models.}
We study state-of-the-art language models, most of which \textit{do not} provide access to internal probabilities as of the writing of this paper --- \textit{GPT-4} ~\citep{openai2023gpt4}, \textit{Claude-v1.3}, and \textit{GPT-3.5-Turbo} ~\citep{openai2022chatgpt}
(June 13th, 2023, snapshot). We also study a few recent models which \textit{do} provide model probabilities for systematic comparisons --- \textit{Llama 2} and \textit{Llama 2 Chat} (70B and 13B sizes) ~\citep{touvron2023llama2} and \textit{text-davinci-003} ~\cite{openai2023text-davinci}. If Llama 2 is mentioned in the text without further identifiers, we refer to the Llama 2 70B base model.

%% file: sections/methods.tex
\subsection{Confidence Elicitation Methods}
\begin{wrapfigure}{r}
{0.58\textwidth}
    \centering
    \vspace{-0.1in}
\includegraphics[width=0.58\textwidth]
{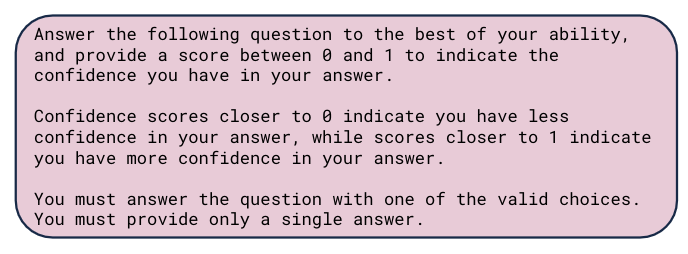}
\centering
\caption{\centering\textbf{Linguistic Confidence Prompt} Instruction for the best linguistic confidence prompt (see exact prompt in Appendix \ref{sec:ling_conf_prompt}).}
\vspace{-0.2in}
\label{fig:prompt_inst}
\end{wrapfigure}

\textbf{Linguistic Confidences.}
For each question, we zero-shot prompt models with an instruction to output a valid answer and a confidence assessment of that answer, sampling the answer and confidence together in a single generation. We generate greedily with temperature $T=0$, and define these confidence estimates generated by the model to be linguistic confidences. Since there can be many ways of eliciting linguistic confidences, we experiment with 24 different prompts across various categories (chain-of-thought, different instructions, fake few shot examples). We find the results to be consistent across prompts, so we report results on our best prompt (see Figure \ref{fig:prompt_inst} for an example instruction of linguistic confidence elicitation).
Section \ref{sec:ling_confs} assesses the quality of linguistic confidences and signals a need for better confidence estimation methods.

\textbf{Model Probabilities.} Models such as Llama 2 and text-davinci-003 provide token-level probabilities for text. We let the confidence score be the probability of the generated answer choice.

\textbf{Surrogate models for confidences.} Since models such as GPT-4 do not give a confidence estimate, we propose using a surrogate model (e.g., Llama 2) to provide confidence estimates. Formally, given an input $x$ we output $y(x) = y_{\textsf{gpt-4}}(x)$ (GPT-4's answer) and 
$\conf(x) = \conf_{\textsf{Llama 2}}(x)$ (Llama 2's confidence in its own answer).
Even though these confidence scores come from a \emph{different} model, Section~\ref{sec:surrogate_confidence_models} shows that the surrogate confidence method outperforms linguistic confidence scores.

\textbf{Mixture of models.} We also propose a mixture of models method where we combine the linguistic confidence from the main model and the surrogate model's confidence score: 
given input $x$ we output $(1 - \alpha) \conf_M(x) + \alpha \conf_S(x)$ where $M$ is the main model and $S$ is the surrogate model. 

We use Llama 2 70B as the surrogate model for all main models since it performs the best. We optimize $\alpha$ to minimize AUC, sweeping over values from 0 to 1. Interestingly, in Section~\ref{sec:mixtures_confidence_models}, we show that even $\alpha=0.001$ works well.

%% file: sections/results/linguistic_confidence_results.tex
\section{Linguistic confidences: asking the model for its confidence}
\label{sec:ling_confs}

As of November 2023, state-of-the-art language models such as GPT-4 and Claude-v1.3 do not give access to internal model probabilities.
In this section, we examine linguistically eliciting confidence: prompt models to assign their answers a confidence score between 0 and 1.
We find that these linguistic confidences leave a lot of room for improvement (around 50-65\% AUROC, compared to 50\% for a random guessing baseline).
These linguistic confidences are also much worse than internal model probabilities when available (for weaker models such as text-davinci-003 and Llama 2). 
We show AUC and AUROC results on all datasets and models in Table~\ref{auroc_linguistic_prob}.

\begin{wrapfigure}{r}{0.485\textwidth} 
\centering
\vspace{-0.26in}
\begin{minipage}{0.485\textwidth}
\begin{algorithm}[H]
\SetKwFunction{SurrogateModel}{SurrogateModel}
\SetKwFunction{MainModel}{MainModel}
\SetKwFunction{append}{append}
\SetKwFunction{AUROC}{AUROC}
\SetKwFunction{AUC}{AUC}
\SetKwFunction{Reward}{Reward}
  \caption{Mixture of Models Confidence}
  \KwData{A question $x$}
  \KwResult{A prediction $\hat{y}$, a confidence score $c$}
    $\hat{y}$, $c_1$ = \MainModel($x$) \;
    $c_2$ = \SurrogateModel($x$) \;
    $c = (1-\alpha) c_1 + \alpha c_2$ \;
\label{mixture-of-models-alg}
\end{algorithm}
\end{minipage}
\vspace{-0.2in}
\end{wrapfigure}

\textbf{Linguistic confidences 
leave room for improvement.}
The AUROC values of linguistic confidences from text-davinci, Llama 2 70b, and GPT-3.5 are close to 50\% (Table~\ref{auroc_linguistic_prob}), which is the score achieved by guessing a random confidence, indicating that linguistic confidences are not a reliable means of separating correct and incorrect examples.
The linguistic confidences of the strongest models, Claude-v1.3 and GPT-4, are better and result in AUROCs in the 60-65\% range, but still leave a lot of room for improvement. 
The AUCs of linguistic confidences
are close to their accuracy (Appendix ~\ref{model_accuracies}) (which is the score achieved by a random guessing baseline) for text-davinci-003 (57.1\% vs 57.7\%), GPT-3.5 (58.1\% vs 59.0\%), and Llama 2 (58.8\% vs 62.4\%). 
Linguistic confidences for the best models are reasonable, but still leave room for improvement ---
GPT-4 has an accuracy of 73.5\% and 
AUC of 80.5\%;
and Claude-v1.3 has an accuracy of 65.5\% and 
AUC of 73.5\%.

\textbf{Linguistic confidences are worse than model probabilities.}
The best current models (GPT-4 and Claude-v1.3) do not provide model probabilities, but we compare the quality of model probabilities and linguistic confidences for text-davinci-003 and the Llama 2 models. For these models, the model probabilities result in better AUC and AUROC values for all of our datasets (Table \ref{auroc_linguistic_prob}). For Llama 2, the model probabilities achieve a \emph{
10.7\% higher AUC
and 19.0\% higher AUROC}
than the linguistic confidences. The Chat model (Llama 2 70B Chat) shows similar trends (Appendix \ref{llama-2-70b-results}).

\textbf{Linguistic confidences are robust to prompt variations.} We examine linguistic confidences using 24 distinct prompts, including asking for numerical confidence or probability scores, asking the model to categorize its confidence into `not sure', `sure', and `very sure', allowing the model to explain confidences with chain-of-thought, asking the model for its confidence in a follow-up question, and varying the prompt instructions. We show results for the best prompt, as there was very little difference in performance across prompts --- our results hold for other prompts as well. A more detailed description of the prompts investigated and the method for selecting the best prompt can be found in Appendix \ref{sec:ling_conf_prompt}.

\input{sections/results/ling_confs_vs_probs_table}
\textbf{Linguistic confidences improve with scale, but not enough.} The quality of linguistic confidences improves with model scale. We see that GPT-4 and Claude-v1.3 have the best linguistic confidences, followed by the Llama 2 70B models, GPT-3.5, and finally text-davinci-003. While the \emph{linguistic confidences} from GPT-4 are not bad (65\% average AUROC), they are worse than \emph{model probabilities} from Llama 2 70b (74\%) and even text-davinci-003 (72\%). Note that AUC scores increase with accuracy --- GPT-4 Linguistic has the highest AUC because GPT-4 has much higher accuracy than Llama 2. The overall utility of a selective classifier depends on both its accuracy and confidence quality, so in the next section we examine ways to improve the confidences of our best-in-class models --- GPT-4 and Claude-v1.3.

%% file: sections/results/ling_confs_vs_probs_table.tex
\begin{table}[t]
\begin{tabular}{@{}lllllllll@{}}
\toprule
 & Confidence Type & TQA & Medqa & CSQA & OBQA & Law & Ethics & Physics \\ \midrule
\multirow{7}{*}{AUC} & Text-davinci Linguistic & 0.523 & 0.504 & 0.718 & 0.775 & 0.532 & 0.590 & 0.579 \\
 & Text-davinci Prob & \textbf{0.607} & \textbf{0.656} & \textbf{0.861} & \textbf{0.929} & \textbf{0.714} & \textbf{0.783} & \textbf{0.697} \\ \cmidrule(l){2-9} 
 & Llama 2 Linguistic & 0.600 & 0.616 & 0.693 & 0.802 & 0.605 & 0.707 & 0.638 \\
 & Llama 2 Prob & \textbf{0.711} & \textbf{0.735} & \textbf{0.804} & \textbf{0.923} & \textbf{0.749} & \textbf{0.834} & \textbf{0.763} \\ \cmidrule(l){2-9} 
 & GPT-3.5 Linguistic & 0.620 & 0.536 & 0.693 & 0.776 & 0.508 & 0.674 & 0.526 \\
 & Claude-v1.3 Linguistic & 0.741 & 0.718 & \textbf{0.807} & 0.879 & 0.669 & \textbf{0.894} & 0.736 \\
 & GPT-4 Linguistic & \textbf{0.889} & \textbf{0.841} & 0.802 & \textbf{0.960} & \textbf{0.732} & 0.869 & \textbf{0.819} \\ \midrule
\multirow{7}{*}{AUROC} & Text-davinci Linguistic & 0.525 & 0.500 & 0.503 & 0.509 & 0.500 & 0.500 & 0.500 \\
 & Text-davinci Prob & \textbf{0.718} & \textbf{0.696} & \textbf{0.806} & \textbf{0.840} & \textbf{0.715} & \textbf{0.758} & \textbf{0.637} \\ \cmidrule(l){2-9} 
 & Llama 2 Linguistic & 0.618 & 0.541 & 0.555 & 0.484 & 0.517 & 0.602 & 0.593 \\
 & Llama 2 Prob & \textbf{0.745} & \textbf{0.722} & \textbf{0.731} & \textbf{0.777} & \textbf{0.733} & \textbf{0.868} & \textbf{0.732} \\ \cmidrule(l){2-9} 
 & GPT-3.5 Linguistic & 0.535 & 0.500 & 0.526 & 0.518 & 0.508 & 0.509 & 0.504 \\
 & Claude-v1.3 Linguistic & \textbf{0.701} & 0.586 & \textbf{0.639} & 0.647 & \textbf{0.586} & \textbf{0.760} & \textbf{0.652} \\
 & GPT-4 Linguistic & 0.665 & \textbf{0.716} & 0.551 & \textbf{0.656} & 0.591 & 0.720 & 0.522 \\ \bottomrule
\end{tabular}
\centering
\begin{tabular}{@{}llllllll@{}}
\toprule
 & Confidence Type & Econ & Algebra & Chem & Security & Policy & Avg \\ \midrule
\multirow{7}{*}{AUC} & Text-davinci Linguistic & 0.412 & 0.300 & 0.440 & 0.690 & 0.856 & 0.577 \\
 & Text-davinci Prob & \textbf{0.431} & \textbf{0.338} & \textbf{0.644} & \textbf{0.891} & \textbf{0.939} & \textbf{0.707} \\ \cmidrule(l){2-8} 
 & Llama 2 Linguistic & 0.415 & 0.189 & 0.474 & 0.817 & 0.930 & 0.624 \\
 & Llama 2 Prob & \textbf{0.498} & \textbf{0.263} & \textbf{0.647} & \textbf{0.866} & \textbf{0.981} & \textbf{0.731} \\ \cmidrule(l){2-8} 
 & GPT-3.5 Linguistic & 0.430 & 0.319 & 0.465 & 0.724 & 0.806 & 0.590 \\
 & Claude-v1.3 Linguistic & 0.640 & 0.333 & 0.653 & 0.812 & 0.934 & 0.735 \\
 & GPT-4 Linguistic & \textbf{0.643} & \textbf{0.551} & \textbf{0.683} & \textbf{0.903} & \textbf{0.965} & \textbf{0.805} \\ \midrule
\multirow{7}{*}{AUROC} & Text-davinci Linguistic & 0.500 & 0.500 & 0.500 & 0.500 & 0.506 & 0.504 \\
 & Text-davinci Prob & \textbf{0.549} & \textbf{0.532} & \textbf{0.695} & \textbf{0.858} & \textbf{0.795} & \textbf{0.717} \\ \cmidrule(l){2-8} 
 & Llama 2 Linguistic & 0.533 & 0.424 & 0.520 & 0.613 & 0.576 & 0.548 \\
 & Llama 2 Prob & \textbf{0.622} & \textbf{0.546} & \textbf{0.732} & \textbf{0.775} & \textbf{0.871} & \textbf{0.738} \\ \cmidrule(l){2-8} 
 & GPT-3.5 Linguistic & 0.518 & 0.522 & 0.505 & 0.519 & 0.519 & 0.515 \\
 & Claude-v1.3 Linguistic & \textbf{0.573} & 0.543 & 0.708 & 0.687 & 0.645 & 0.644 \\
 & GPT-4 Linguistic & 0.551 & \textbf{0.599} & \textbf{0.721} & \textbf{0.750} & \textbf{0.753} & \textbf{0.650} \\ \bottomrule
\end{tabular}
\caption{\textbf{AUC and AUROC - Linguistic Confidences vs Model Probabilities} We compare the AUC and AUROC values for linguistic confidences and model probabilities in weaker models (text-davinci-003 and Llama 2 70B), and find that model probabilities consistently outperform linguistic confidences. For closed source models (which don't provide model probabilities), we see that Claude-v1.3 and GPT-4 provide the best linguistic confidences in both AUC and AUROC.
}
\label{auroc_linguistic_prob}
\end{table}

%% file: sections/results/surrogate_mixture_confidence_results.tex
\section{Surrogate models are reliable confidence estimators}
\label{sec:surrogate_confidence_models}
In the previous section we found that linguistic confidences 
leave room for improvement.
Here we show that model probabilities from a separate `surrogate' model can surprisingly provide better confidence estimates for a model than its own linguistic confidence scores, even though the probabilities come from a different model.
\input{sections/results/auc_surrogate_v_main_heatmap}
\subsection{Results}
\textbf{Surrogate model confidences outperform linguistic confidences.}
AUC improves for all models when probabilities from
 a
surrogate model are used, as opposed to using the model's own linguistic confidences.
Figure \ref{table:surrogate_auc_heatmap} shows a heatmap of the AUC for different main models (on the $x$-axis) as we vary the surrogate model (on the $y$-axis).
We see that model probabilities (bottom
four
rows) lead to higher AUC (are darker) than linguistic confidences (top
 six
rows)
even when the probabilities come from a different model.
For example, using Llama 2 70B probabilities as a surrogate improves AUC from 
80.5\% to 82.1\%
for GPT-4, 
73.5\% to 76.3\%
for Claude-v1.3, and 
59.0\% to 72.1\%
for GPT-3.5, and AUROC also shows similar increases for all models (Table \ref{surrogate_mix_all_models}, Figure \ref{fig:selective_acc_gpt4}).

\textbf{Weak surrogates are also good confidence estimators.}
Even using Llama 2 13B or text-davinci-003 as a surrogate leads to comparable or better performance than using a model's own linguistic confidences.
We found this intriguing because these models are much smaller and less accurate, e.g., Llama 2 13B has an average accuracy of 47.2\% vs. 65.5\% for Claude-v1.3 and 73.5\% for GPT-4.

\textbf{Other findings.}
Recent work suggests chat models trained using reinforcement learning from human feedback (RLHF)
might be less calibrated than base models. In Appendix \ref{sec:surrogate_model_details}, we compare chat and base model probabilities as surrogate confidences and find that Llama 2 70B base slightly outperforms Llama 2 70B chat in selective classification with both linguistic confidences and model probabilities --- but both models perform similarly as surrogates. As we might expect, in general better models (such as Llama 2 70B) are better surrogates.
Finally, we find that \emph{linguistic confidences} from stronger models can provide good surrogate confidences for weaker models --- the AUC of 
GPT-3.5 improves by 5.7\%
when using GPT-4's linguistic confidences instead of its own.

\section{Mixtures of models for better confidence estimates}
\label{sec:mixtures_confidence_models}
In the previous section, we proposed the use of surrogate models --- using a main model to produce answers and a separate, surrogate to estimate the main model's confidence in the answers --- and found surrogates to outperform linguistic confidence scores elicited from the main model.
In this section, we find that 
the signals from linguistic confidences and surrogate probabilities are complementary
--- the two can be composed
to get state of the art confidence estimates for all models.
 
\subsection{Results}
\begin{figure}[t]
    \begin{subfigure}{0.5\textwidth} 
        \centering
        \includegraphics[width=\linewidth]{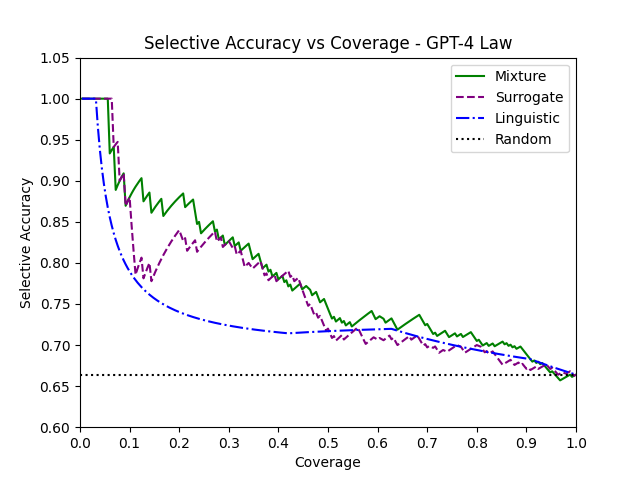} 
        \caption{MMLU - Professional Law}
        \label{fig:subfig1}
    \end{subfigure}%
    \begin{subfigure}{0.5\textwidth}
        \centering
        \includegraphics[width=\linewidth]{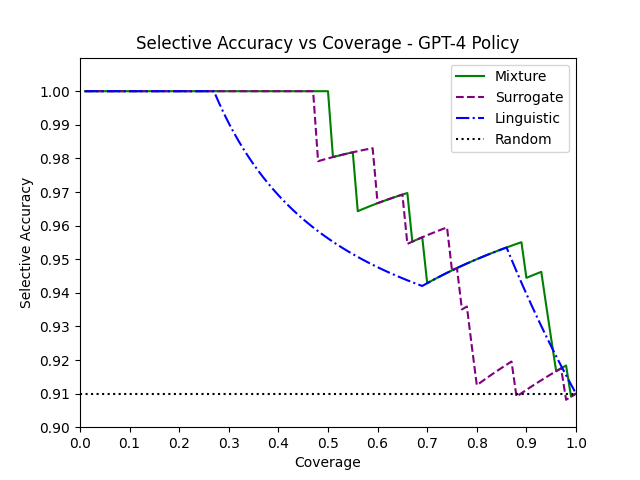} 
        \caption{MMLU - US Foreign Policy}
        \label{fig:subfig2}
    \end{subfigure}
    \caption{\textbf{Selective Accuracy vs. Coverage for GPT-4.} Our surrogate and mixture methods have a higher area under the selective accuracy vs coverage curve (AUC) than the linguistic confidence and random confidence baselines. We plot the coverage $c$ on the $x$-axis and the selective accuracy (accuracy on the top $c$ fraction of examples) on the $y$-axis, for two representative tasks. Notice that the mixture (green solid) and surrogate (purple dashed) lines are above the linguistic confidence (blue dashed/dotted) and random guessing baseline (black dotted).}
    \label{fig:selective_acc_gpt4}
\end{figure}

\input{sections/results/compare_methods_all_models_table}
\textbf{Mixtures of models provide best confidences.} Mixing surrogate and linguistic confidences (Algorithm ~\ref{mixture-of-models-alg}) leads to the best confidence estimates for all models --- AUCs increase from 
80.5\% to 83.4\%
for GPT-4 and 
73.5\% to 77.2\%
for Claude-v1.3 (Table ~\ref{surrogate_mix_all_models}). The optimal $\alpha$ (Algorithm ~\ref{mixture-of-models-alg}) for best average performance across tasks is $0.4$ for GPT-4 and $0.6$ for Claude-v1.3. AUROCs also increase for these models, by 5.3\%
for GPT-4 and 5.0\% 
for Claude-v1.3 (Table \ref{surrogate_mix_all_models}). We also plot the selective accuracy against coverage in Figure \ref{fig:selective_acc_gpt4}, where the mixture and surrogate method lie above the linguistic confidences curve.

\textbf{Epsilon is all you need.}
\label{sec:tiebreaking}
We also study a special case of mixtures called tiebreaking, where we set $\alpha$ to a small value $\epsilon \rightarrow 0$ (Algorithm ~\ref{mixture-of-models-alg}) --- this simply uses the surrogate model to `break ties' and provide relative ordering for examples with the same linguistic confidence.
Adding only 0.1\% of a surrogate model's probabilities to a model's linguistic confidences performs better than using either the linguistic confidences or surrogate probabilities alone, and closely matches performance of the optimal $\alpha$ (Table \ref{surrogate_mix_all_models}).
For GPT-4, tiebreaking achieves
86\%
of the AUC gains (over linguistic confidences) of the optimal $\alpha$, and 87\% of the AUROC gains.

\textbf{Mixing surrogate and self-consistency confidences leads to further gains.} 
Concurrent work~\citep{xiong2023can} on eliciting linguistic confidences uses self-consistency (SC) to sample multiple linguistic confidence scores for each answer and aggregates them through a post processing technique.
For further gains, we experiment with leveraging these SC-based linguistic confidences for GPT-4 --- we replace linguistic confidences $c_1$ in Algorithm~\ref{mixture-of-models-alg}
with the outputs of their best method (hybrid self-consistency).
The updated Algorithm ~\ref{mixture-of-models-alg}
leads to state-of-the-art confidence estimates, also outperforming their hybrid self-consistency technique (Table~\ref{gpt4_all_methods}), with an overall 
4.1\%
gain in AUC for GPT-4 over vanilla linguistic confidences, and a 
9.1\%
gain in AUROC.

\textbf{Other findings.} Probabilities of smaller surrogate models can also be composed with linguistic confidences --- composing Llama 2 13B's probabilities with GPT-4's linguistic confidences retains
66\%
of the AUC gains seen from composing GPT-4 with Llama 2 70B. Composing GPT-4 and Claude-v1.3's linguistic confidences can boost GPT-4's AUC by 2.1\% and AUROC by 3\%,
indicating that linguistic confidences of different models can provide complementary estimates of uncertainty.
Additionally, we find that even composing the model probabilities of two different models can provide better confidence estimates --- composing Llama 2's probabilities with those of Llama 2 Chat improves Llama 2's AUC from 73.1\% to 73.8\% and AUROC from 73.8\% to 74.5\%. Mixing confidences from more than two models could potentially lead to further improvements.
\input{sections/results/gpt4_compare_methods_table}

%% file: sections/results/auc_surrogate_v_main_heatmap.tex
\begin{figure}
\centering
    \includegraphics[scale=0.52]
{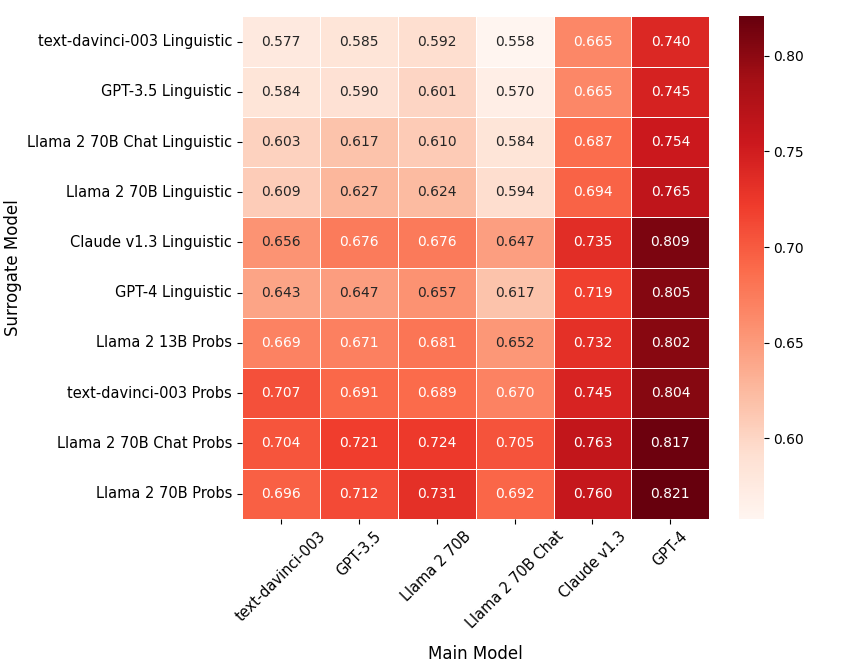}
    \caption{\textbf{AUCs for Different Surrogate Models.} We plot the AUC as we vary the main model (on the $x$-axis) and the surrogate model (on the $y$-axis). Using surrogate model probabilities as confidence estimates improves AUCs for all models over their own linguistic confidences---the bottom 
    4
    rows (surrogate probabilities) are darker than the top
    6
    rows (linguistic confidences). Even model probabilities from a smaller 
    Llama 2 13B model 
    lead to comparable or better AUCs
    for all models.}
    \label{table:surrogate_auc_heatmap}
\end{figure}

%% file: sections/results/compare_methods_all_models_table.tex
\begin{table}[t]
\centering
\begin{tabular}{@{}llllllll@{}}
\toprule
                       &                        & \begin{tabular}[c]{@{}l@{}}Text-davinci\end{tabular} & \begin{tabular}[c]{@{}l@{}}GPT-3.5 \end{tabular} & \begin{tabular}[c]{@{}l@{}}Llama 2\end{tabular} & Claude-v1.3     & GPT-4           \\ \midrule
\rowcolor{grayrowcolor}
\multirow{4}{*}{AUC}   & Ling. Conf. & 0.577 & 0.590  & 0.624  & 0.735 & 0.805 \\ 
 & Surrogate$^\dagger$  & 0.707 & 0.719 & \textbf{0.731}  & 0.763 & 0.821 \\  
 & Tiebreak$^\dagger$ & \textbf{0.711} & 0.719  & 0.715  & 0.764  & 0.830 \\  
 & Mixture of Models$^\dagger$ &\textbf{0.711} & \textbf{0.722} & \textbf{0.731} & \textbf{0.772} & \textbf{0.834} \\
 \midrule
\rowcolor{grayrowcolor}
\multirow{4}{*}{AUROC} & Ling. Conf. & 0.504   & 0.514 & 0.548 & 0.637 & 0.646  \\ 
& Surrogate$^\dagger$ & 0.717  & 0.708  & \textbf{0.738} & 0.671 & 0.657 \\ 
& Tiebreak$^\dagger$ & \textbf{0.718}  & 0.708  & 0.699  & 0.683 & 0.692 \\  
 & Mixture of Models$^\dagger$ & \textbf{0.718}  & \textbf{0.709} & 0.737 & \textbf{0.687} & \textbf{0.699} \\ \bottomrule
\end{tabular}
\caption{\textbf{AUC and AUROC of Surrogate and Mixture of Model Methods.} We compare the performance of our proposed methods$^\dagger$ with the baseline linguistic confidence method (gray). For both AUC and AUROC, our proposed methods outperform linguistic confidences on all models. Mixture of models improves the AUC of GPT-4 by 
3\%
and AUROC by 5\%.}
\label{surrogate_mix_all_models}
\end{table}

%% file: sections/results/gpt4_compare_methods_table.tex
\begin{table}[t]
\begin{tabular}{@{}lllllllll@{}}
\toprule
 & Method & TQA & Medqa & CSQA & OBQA & Law & Ethics & Physics \\ \midrule
 \rowcolor{grayrowcolor}
\multirow{6}{*}{AUC} & Ling. Conf. & 0.889 & 0.841 & 0.802 & 0.960 & 0.732 & 0.869 & 0.819 \\
\rowcolor{grayrowcolor}
 & SC Ling. Conf. & 0.903 & \textbf{0.887} & 0.841 & 0.978 & 0.729 & \textbf{0.902} & 0.846 \\
 & Surrogate$^\dagger$ & 0.866 & 0.844 & 0.849 & 0.965 & 0.762 & 0.849 & \textbf{0.891} \\
 & Tiebreak$^\dagger$ & 0.902 & 0.871 & 0.833 & 0.967 & 0.768 & 0.889 & 0.861 \\
 & Mixture$^\dagger$ & 0.895 & 0.864 & 0.849 & 0.969 & \textbf{0.780} & 0.882 & 0.886 \\
 & SC Mixture$^\dagger$ & \textbf{0.921} & 0.873 & \textbf{0.877} & \textbf{0.979} & 0.757 & 0.894 & 0.881 \\ \midrule
 \rowcolor{grayrowcolor}
\multirow{6}{*}{AUROC} & Ling. Conf. & 0.665 & 0.716 & 0.551 & 0.656 & 0.591 & 0.720 & 0.522 \\
\rowcolor{grayrowcolor}
 & SC Ling. Conf. & 0.698 & \textbf{0.767} & 0.625 & 0.833 & 0.619 & \textbf{0.817} & 0.592 \\
 & Surrogate$^\dagger$ & 0.543 & 0.666 & 0.656 & 0.683 & 0.619 & 0.617 & 0.648 \\
 & Tiebreak$^\dagger$ & 0.671 & 0.750 & 0.611 & 0.716 & 0.628 & 0.740 & 0.589 \\
 & Mixture$^\dagger$ & 0.642 & 0.731 & 0.646 & 0.731 & 0.655 & 0.711 & 0.648 \\
 & SC Mixture$^\dagger$ & \textbf{0.702} & 0.747 & \textbf{0.679} & \textbf{0.838} & \textbf{0.655} & 0.783 & \textbf{0.663} \\ \bottomrule
\end{tabular}
\centering
\begin{tabular}{@{}llllllll@{}}
\toprule
 & Method & Econ & Algebra & Chem & Security & Policy & \textbf{Avg} \\ \midrule
 \rowcolor{grayrowcolor}
\multirow{6}{*}{AUC} & Ling. Conf. & 0.643 & 0.551 & 0.683 & 0.903 & 0.965 & 0.805 \\
\rowcolor{grayrowcolor}
 & SC Ling. Conf. & 0.663 & 0.584 & 0.726 & 0.915 & 0.965 & 0.828 \\
 & Surrogate$^\dagger$ & \textbf{0.667} & 0.572 & 0.724 & 0.888 & 0.971 & 0.821 \\
 & Tiebreak$^\dagger$ & 0.654 & 0.580 & 0.746 & 0.910 & 0.974 & 0.830 \\
 & Mixture$^\dagger$ & 0.664 & 0.581 & 0.749 & 0.908 & \textbf{0.976} & 0.834 \\
 & SC Mixture$^\dagger$ & 0.662 & \textbf{0.645} & \textbf{0.763} & \textbf{0.926} & 0.973 & \textbf{0.846} \\ \midrule
 \rowcolor{grayrowcolor}
\multirow{6}{*}{AUROC} & Ling. Conf. & 0.551 & 0.599 & 0.721 & 0.750 & 0.753 & 0.650 \\
\rowcolor{grayrowcolor}
 & SC Ling. Conf. & 0.622 & 0.682 & 0.818 & 0.798 & 0.755 & 0.719 \\
 & Surrogate$^\dagger$ & 0.578 & 0.621 & 0.706 & 0.779 & 0.764 & 0.657 \\
 & Tiebreak$^\dagger$ & 0.569 & 0.648 & 0.760 & 0.815 & 0.805 & 0.692 \\
 & Mixture$^\dagger$ & 0.578 & 0.648 & 0.759 & 0.814 & \textbf{0.822} & 0.699 \\
 & SC Mixture$^\dagger$ & \textbf{0.595} & \textbf{0.763} & \textbf{0.819} & \textbf{0.839} & 0.810 & \textbf{0.741} \\ \bottomrule
\end{tabular}
\caption{\textbf{AUC and AUROC of All Confidence Methods for GPT-4.} 
Our proposed surrogate model method outperforms linguistic confidences on
9/12 datasets on AUC.
Mixing surrogate probabilities and linguistic confidences outperforms vanilla linguistic confidences on AUC for all 12 datasets.
The mixture of surrogate probabilities also outperforms hybrid self-consistency confidences, the best method in~\citet{xiong2023can}, on average (AUC 
83.4\% vs 82.8\%.
Mixing surrogate probabilities with self-consistency linguistic confidences leads to the best confidence estimates overall, outperforming all methods with an average 
84.6\%
AUC and 
74.1\%
AUROC, which is a gain of 
4.1\%
and
9.1\%
respectively over vanilla linguistic confidences.}
\label{gpt4_all_methods}
\end{table}

%% file: sections/results/analysis.tex
\section{Analysis}
\label{sec:analysis}
\textbf{Why Are Vanilla Linguistic Confidences Worse Than Model Probabilities?} 
In Section \ref{sec:ling_confs}, we showed that linguistic confidences underperformed model probabilities. Here we provide some intuitions for this behavior. We observe that the distribution of model probabilities is quite varied (1456 unique values for Llama 2 70B across 12 datasets), while the distribution of linguistic confidences is quite clustered (only 8 unique values for GPT-4 across 12 datasets). This clustering may be because training corpora contain higher frequencies of ``nice'' probability numbers such as 90\% or 100\%~\citep{zhou2023navigating}. The repetitiveness of linguistic confidences, compared to model probabilities, hinders relative confidence ordering and good AUC and AUROC performance --- GPT-4 repetitively generates 0.9 for 50\% of examples across 12 tasks, so it cannot separate them. We tried simple ablations to increase linguistic confidence variation, by increasing the temperature of generations or instructing the model `It's ok to be less sure of your answers.', but they did not improve AUC because they reduced model accuracy.

\begin{figure}[t]
\centering
    \begin{subfigure}{0.45\textwidth}
        \centering
        \includegraphics[width=\linewidth]{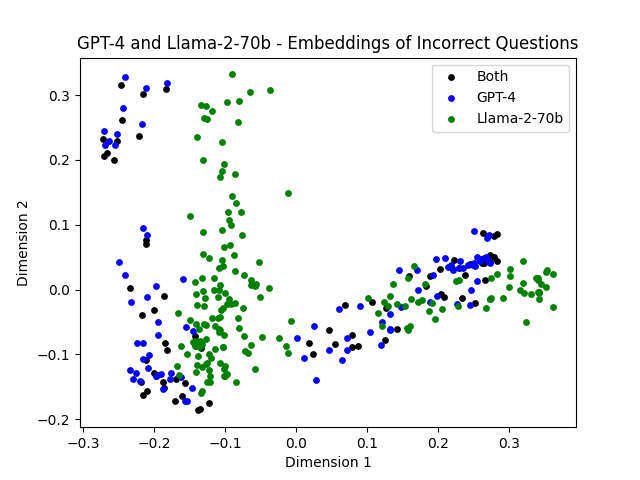} 
        \caption{GPT-4 and Llama 2 70B}
        \label{fig:llama2_70b_embed}
    \end{subfigure}%
    \begin{subfigure}{0.45\textwidth}
        \centering
        \includegraphics[width=\linewidth]{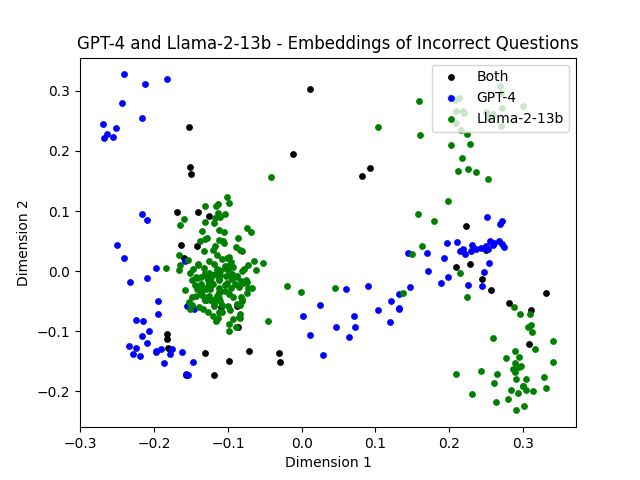} 
        \caption{GPT-4 and Llama 2 13B}
        \label{fig:llama2_13b_embed}
    \end{subfigure}
    \caption{\textbf{Embeddings of Incorrect Questions for GPT-4 and Surrogate Models} Plots of the embeddings of questions GPT-4 and two surrogate models (Llama 2 70B and Llama 2 13B) answer incorrectly on two representative datasets - TruthfulQA and College Chemistry. Questions only GPT-4 answers incorrectly are in blue, questions GPT-4 and the surrogate answer incorrectly are in black, and questions only the surrogate answers incorrectly are in green. There are more questions that both GPT-4 and Llama 2 70B answer incorrectly and more semantic similarity in their incorrect questions. This indicates that Llama 2 70B and GPT-4 struggle with semantically related concepts and that the 70B model may more closely
    estimate
    GPT-4's uncertainty than the 13B model.}
    \label{fig:gpt3_llama_embeddings}
\end{figure}

\textbf{Why Does Surrogate Confidence Estimation Work?} In Section \ref{sec:surrogate_confidence_models}, we demonstrate that models can receive good quality confidence estimates from other surrogate models. In this section, we provide some intuitions for our results. We find that for a main model $M$, a model $S$ tends to be a better surrogate when there is a higher correlation in the questions answered correctly by $M$ and $S$. The questions GPT-4 answers correctly are more correlated with those that Llama 2 70B answers correctly (Pearson correlation of 0.39), than those that Llama 2 13B answers correctly (correlation 0.19) (Appendix \ref{sec:analysis_appendix}). We also plot the embeddings of questions that GPT-4 gets incorrect (blue dots) and the questions two potential surrogates Llama 2 70B and Llama 2 13B get incorrect (green dots) (Figure \ref{fig:gpt3_llama_embeddings}). GPT-4 and Llama 2 70B tend to make mistakes on more of the same questions (more black dots on the left plot). We also see more spatial similarity in the mistakes of GPT-4 and Llama 2 70B. So better surrogate models $S$ and their corresponding main models $M$ may struggle with \textit{semantically related} concepts, causing them to have low confidences on similar types of questions. Intuitively, the probabilities of a surrogate like Llama 2 transfer well to a stronger model like GPT-4 because Llama 2 is good at `spotting' difficult questions, even if it cannot always answer them --- we reason that both models have higher entropy probability distributions over answer choices for more difficult questions, and more peaked probability distributions for easier questions.

\textbf{Why Is Tiebreaking Sufficient?} As mentioned, linguistic confidences tend to be repetitive and clustered at only a few values (e.g., 0.9), limiting their ability to separate correct and incorrect answers. Since a surrogate model's probabilities for each example are nearly unique, composing just a small fraction of them with linguistic confidence scores (Section \ref{sec:tiebreaking}) can allow answers which previously had the same linguistic confidence to now be separable through different composite confidence scores. This means that in cases where linguistic confidence scores are identical, we fall back on the surrogate model's probabilities to provide an order examples based on confidence.

%% file: sections/related_work.tex
\section{Related Work}
\textbf{Confidence Estimation for LLMs.} Confidence estimation for LLMs has been studied in several related works.~\citet{kadavath2022language} show that Claude’s model probabilities are well-calibrated on multiple/choice and True/False questions.~\citet{zhou2023navigating} study the effect of introducing expressions of uncertainty into prompts, on model accuracy. Our work differs from these since we focus on confidence elicitation for models which don’t provide log probabilities. Concurrent work~\citep{xiong2023can} studies calibration and selective classification of linguistic confidence scores generated by LLMs. While this work also elicits prompted confidences, they focus on self-consistency (SC) based methods which are expensive because they require prompting GPT-4 several times. Our proposed Surrogate and Mixture of models methods are less expensive, since model probabilities from smaller models (Llama 2) are used to improve the confidence estimates of larger models (GPT-4). We also show performance improvements over their best method.
~\citep{lin2022teaching} examine fine-tuning language models to improve confidence estimation, which we do not have access to.

\textbf{Selective Classification and OOD Detection.}
Our paper focuses on selective classification, a classical problem in machine learning~\citep{elyaniv2010foundations,khani2016unanimity,feng2019selective,jones2021selective} and statistics~\citep{chow1970optimum,hellman1970probability}.
A related problem is out-of-distribution detection~\citep{pimentel2014review,liang2018enhancing,ovadia2019uncertainty}, where the goal is to detect examples very different from training (where the model may make mistakes).
Prior work uses internals of the models --- probability outputs~\citep{hendrycks2017baseline}, representations~\citep{lee2018unified} of models, or even updates the training procedure~\citep{bartlett2008classification,mozannar2020consistent} --- which state-of-the-art LLMs do not currently give access to.

\textbf{Calibration.}
The general idea of confidence estimation is also studied in calibration~\citep{murphy1977reliability,degroot1983forecasters,naeini2014binary,guo2017calibration}.
While related, the focus is different---a model which outputs its accuracy on every example has 0 calibration error (ECE), but cannot \emph{separate} correct and incorrect examples~\citep{kuleshov2015calibrated}.

%% file: sections/conclusion.tex
\section{Conclusion and Future Work} Our work aims to address the open challenge of eliciting good confidence estimates from state-of-the-art LLMs such as GPT-4 and Claude-v1.3, which don't provide access to their internal probabilities. Our results highlight the importance of releasing model probabilities, since linguistic confidences alone are generally not expressive enough to provide high quality confidence estimates. We demonstrate that
probabilities from weaker white-box, surrogate models can effectively estimate the internal confidences of stronger black-box models like GPT-4, outperforming linguistic confidences, and provide some intuitions for why confidences can transfer between models. We hope that our findings can inspire future work on understanding the transferability of model probabilities and representations and on leveraging this transferability to use white-box models to understand black-box models. Interestingly, we also find that confidence signals from different models are complementary and can be composed for even more reliable confidence estimation. Future methods could further build on this result to develop more sophisticated methods of confidence signal composition. 

%% file: sections/appendix.tex
\appendix
\section{Appendix}
\subsection{Dataset Details}
\label{sec:dataset_details}
\textbf{TruthfulQA} is a multiple choice benchmark designed to check the truthfulness of large language models by testing them on questions across 38 different categories like health and politics where humans might provide incorrect responses due to implicit biases or incorrect beliefs. This task is challenging for language models because they may imbibe these same misconceptions from training corpora.\\\\
\textbf{MedQA} is a challenging dataset testing medical knowledge with questions based on the United States Medical License Exams (USMLE) and other medical board exams.\\\\
\textbf{CommonsenseQA} is a multiple choice benchmark testing commonsense reasoning, with challenging associations extracted from ConceptNet to find many target concepts for a single source concept.\\\\
\textbf{OpenbookQA} is a multiple choice dataset requiring multi-step reasoning over common and commonsense knowledge requiring deeper understanding of a diverse set of topics.\\\\
\textbf{MMLU} is a massive benchmark covering 57 subjects from a diverse set of areas including STEM, humanties, and social sciences. This benchmark tests both more rudimentary and more advanced sets of knowledge for these topics covering great breadth and depth. 

Our method requires no training or adaptation of the models used. We evaluate on 250 examples from each dataset or on the maximum size of the dataset's test subset (if test subset is smaller).
\subsection{Model Accuracies}
\label{model_accuracies}
Following are the accuracies for each of the models on the 12 datasets. Answers are elicited using the prompt format specified in \ref{sec:ling_conf_prompt}. As expected the GPT-4 model has the highest accuracies on all 12 datasets, followed by Claude-v1.3. Llama 2 Chat and Base have comparable accuracy to GPT-3.5. Text-davinci-003 has the lowest accuracies.
\begin{table}[H]
\begin{tabular}{@{}llllllll@{}}
\toprule
Model & TQA & MedQA & CSQA & OBQA & Law & Ethics & Physics \\ \midrule
Text-davinci & 0.472 & 0.504 & 0.712 & 0.772 & 0.532 & 0.590 & 0.579 \\
Llama-2 & 0.524 & 0.564 & 0.664 & 0.808 & 0.572 & 0.590 & 0.583 \\
Llama-2 Chat & 0.480 & 0.512 & 0.684 & 0.728 & 0.528 & 0.600 & 0.528 \\
GPT-3.5 & 0.572 & 0.536 & 0.676 & 0.776 & 0.504 & 0.670 & 0.523 \\
Claude-v1.3 & 0.596 & 0.640 & 0.736 & 0.832 & 0.604 & 0.760 & 0.638 \\
GPT-4 & \textbf{0.836} & \textbf{0.736} & \textbf{0.768} & \textbf{0.940} & \textbf{0.664} & \textbf{0.760} & \textbf{0.813} \\ \bottomrule
\end{tabular}
\centering
\begin{tabular}{@{}lllllll@{}}
\toprule
 & Econ & Algebra & Chem & Security & Policy & Avg \\ \midrule
Text-davinci & 0.412 & 0.300 & 0.440 & 0.690 & 0.850 & 0.571 \\
Llama-2 & 0.386 & 0.240 & 0.460 & 0.750 & 0.910 & 0.588 \\
Llama-2 & 0.333 & 0.310 & 0.420 & 0.670 & 0.850 & 0.554 \\
GPT-3.5 & 0.404 & 0.320 & 0.460 & 0.730 & 0.800 & 0.581 \\
Claude-v1.3 & 0.579 & 0.330 & 0.500 & 0.760 & 0.880 & 0.655 \\
GPT-4 & \textbf{0.596} & \textbf{0.480} & \textbf{0.520} & \textbf{0.800} & \textbf{0.910} & \textbf{0.735} \\ \bottomrule
\end{tabular}
\caption{\textbf{Model Accuracies} Accuracies of all 6 models on all 12 tasks. GPT-4 is the highest performing model for all tasks.}
\end{table}
\vspace{-0.1in}
\subsection{AUC and AUROC Definitions}
\label{sec:auc_auroc_definitions}
\textbf{AUC with Randomized or Deterministic Classifiers.} To plot the accuracy-coverage curve we compute $A(c)$, the selective accuracy at coverage $c$ across different values of $c$. $A(c)$ is the accuracy if the model only makes a prediction on the $c$ proportion of  data with highest confidence scores. for different values of $c$. When making a prediction on $c$ proportion of data, for each example $x$ we use a binary classifier on the confidence score $C(x)$ decide if we are making a prediction for $x$ ($1$ if making a prediction and $0$ if abstaining from prediction). Such a classifier can either be deterministic or randomized.\\\\
\textit{Deterministic Classifiers.} A deterministic classifier $f$ returns identical outputs for identical inputs --- resulting in consistent treatment of examples with the same confidence score (either predict on all or abstain on all).
 Using a deterministic classifier to select $c$ portion of examples to predict on means we find the highest confidence threshold $t$ such that $P(C(x) \geq t) \geq c$ --- $t$ is the highest confidence threshold where the proportion of examples with confidence greater than or equal to $t$ is greater than or equal to the required coverage $c$. With a deterministic classifier, we predict on $P(C(x) \geq t)$ proportion of examples, which may be greater than the required coverage $c$.
\begin{equation}
    f(C(x)) \in \{0, 1\}
\end{equation}
\textit{Randomized Classifiers.} A randomized classifier $h$ can return different outputs for the same input.
Since models can output the same linguistic confidences for multiple examples, a randomized classifier can allow us to achieve exactly a coverage of $c$ by making predictions on some examples with a given confidence, and abstaining on other examples with the same confidence.
To enable $h$ to break ties and make different predictions for examples with the same confidence score,
we add a small amount of Gaussian noise to each confidence score $\mathcal{N}(0, \epsilon), \epsilon \to 0$ to enforce a confidence-based ordering of examples.
\begin{equation}
    h(C(x) + \mathcal{N}(0, \epsilon)) \in \{0, 1\}
\end{equation}
\textit{Deterministic vs Randomized AUC Example.} Suppose a model assigns half of the examples a confidence of 1 and gets them all right, and the other half of examples a confidence of 0.5 and gets 50\% of them right. What is the selective accuracy at coverage 75\%? A deterministic classifier would select 0.5 as $t$ and predict on all examples with $C(x) \geq t$, which in this case is all of the examples (notably leading to a coverage of 100\% instead of 75\%). This would lead to an accuracy of 75\%. A randomized classifier would predict on all examples of confidence 1, but to meet the 75\% coverage threshold, it would predict on half of the examples which have confidence 0.5 --- selecting the top half after adding random noise. This would lead to an accuracy of approximately 83\%. \\\\
The above example demonstrates that we may underestimate the AUC value by using a deterministic classifier - since it forces a prediction on a higher coverage threshold. For example, for GPT-4 using a deterministic classifier leads to 78.8\% AUC averaged across 12 datasets, while using a randomized classifier leads to 80.5\% AUC.
Since it generally leads to higher selective accuracy values and allows predicting on exactly the top $c$ proportion of examples based on confidence, we use a randomized classifier in our AUC calculations.\\\\
\textbf{AUROC Definition.} Following is how we compute the AUROC metric. Let $\correct(x, y) = 1$ if an answer $y$ is correct for input $x$, and $0$ otherwise.
$\conf(x) \in [0, 1]$ is the confidence score for example $x$.:\\
\begin{enumerate}
    \item Let the true positive rate at threshold $t$ be the fraction of correct examples with confidence at least $t$: $\mbox{TPR}(t) = \E[ \correct(x, y(x)) \mathbb{I}(\conf(x) \geq t) ] / E[ \correct(x, y(x)) ]$
    \item Let the false positive rate at threshold $t$ be the fraction of incorrect examples with confidence at least $t$:
    $\mbox{FPR}(t) =  E[ (1 - \correct(x, y(x))) \mathbb{I}(\conf(x) \geq t) ] / E[ 1 - \correct(x, y(x)) ]$\\
    \item Plot $\mbox{TPR}(t)$ (y-axis) against $\mbox{FPR(t)}$ (x-axis) and take area under the curve.
    The AUROC checks how well thresholding on the confidence can predict if an example is correct or incorrect.
\end{enumerate}
A random classifier (which assigns random confidences to each example) will have the same TPR and FPR.
A model with good confidences will assign higher confidences to correct examples, so a larger fraction of the correct examples (than incorrect examples) will have confidence $\geq t$.

\subsection{Linguistic Confidence Prompts}
\label{sec:ling_conf_prompt} We zero-shot prompt models to elicit answers and linguistic confidence scores. We focus on zero-shot prompting instead of few-shot prompting to assess how good models are at linguistically understanding a notion of confidence, without being primed with examples which could bias their generated confidence assessments.\\\\
\textbf{Prompt Categories.} We study 24 prompt format variations to elicit linguistic confidences, since there can be many ways for a model to describe its confidence levels and we want to ensure a fair assessment of linguistically generated confidence estimates that is not overly conditioned on a specific prompt. These fall into the following categories --- prompts eliciting:
\begin{enumerate}
    \item Numerical confidences (score from between varying ranges 0-1, 0-10, 0-100), probabilities from 0-100\%
    \item Linguistic categorization of confidences into varying numbers of confidence buckets (`not sure', `sure', `very sure', among other phrases and degrees of certainty)
    \item Zero-shot chain-of-thought explanations of confidence in addition to confidence scores
    \item Prompting first for an answer and then re-prompting for the confidence in the answer
    \item Varying the confidence instructions (with other phrases describing confidence including `uncertainty', `certainty', `correctness', `probability or likelihood of correctness')
\end{enumerate}
\textbf{Best Prompt Selection.} We measure the AUC performance of each prompt across the 12 datasets for all models. The optimal prompt varies slightly for each model, so for the best prompt we select the prompt which reduces the sum of drops in AUC per model from the optimal prompt for each individual model (the prompt which is the closest in optimal AUC performance for all models on all datasets). This prompt is used for all of our experiments.

\textbf{Best Prompt Description.} The prompt elicits a confidence score from 0-1 with the given instruction. It also includes \textit{`fake few-shot examples'} as described which show fake questions, answers, and confidence scores to allow the model to learn the format of the task without providing any other task-specific information. Since fake-fewshot examples do including answer choices (D, A in the example) and specific confidence scores (0.4, 0.7), we experimented with variations of the prompt modifying the exact answers and confidences included in the example and found the change in generated confidences to be minimal -- indicating that models were able to use these examples to learn the format of the task without over-conditioning on the specific answers and confidences provided.:
\begin{figure}[H]
\centering
\includegraphics[scale=0.45]{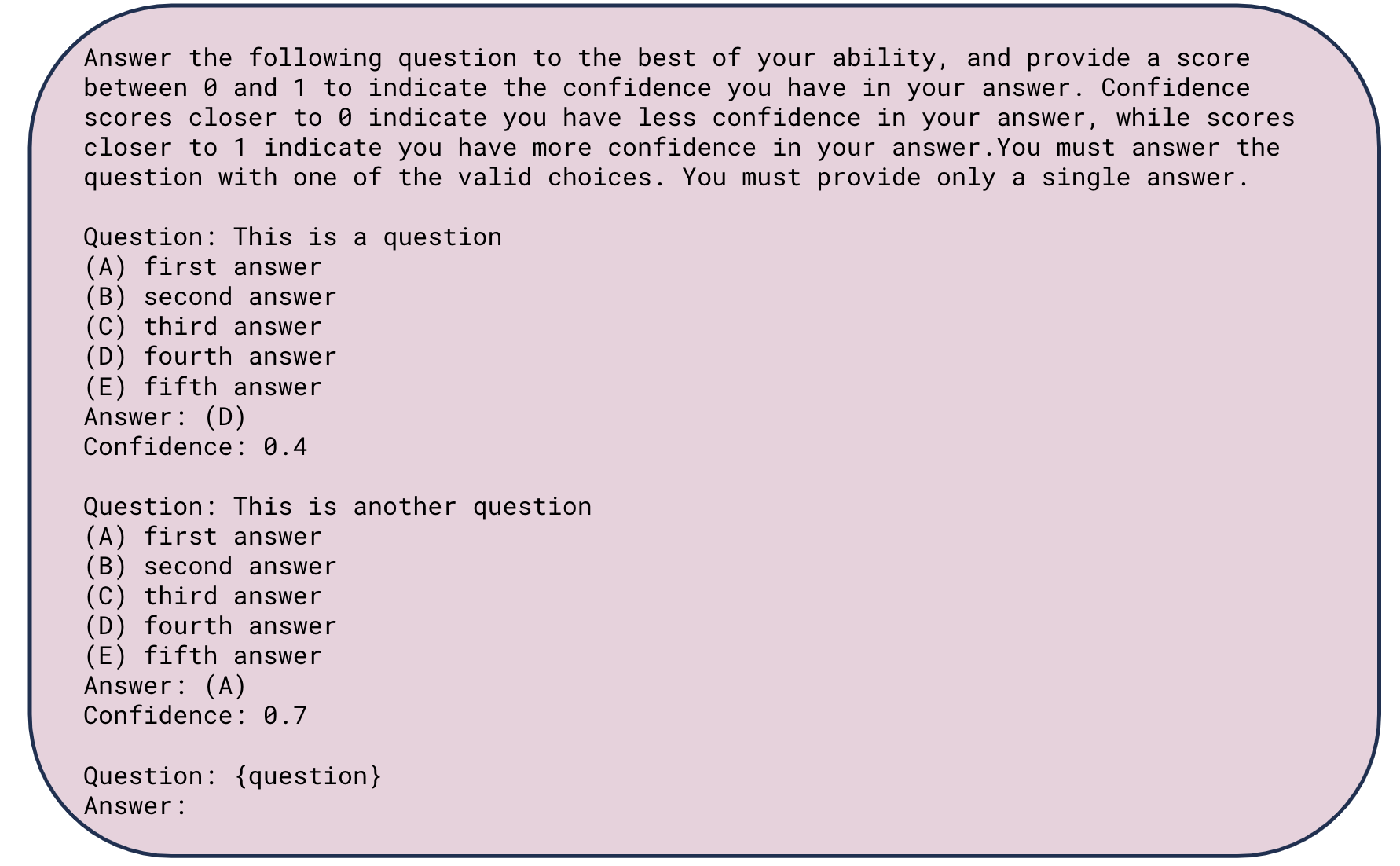}
    \caption{\textbf{Best Linguistic Confidence Prompt}}
\end{figure}
\textbf{Examples of Other Prompts.} Following are a few of the several alternative prompts we tried:
\begin{figure}[H]
\centering
\includegraphics[scale=0.45]{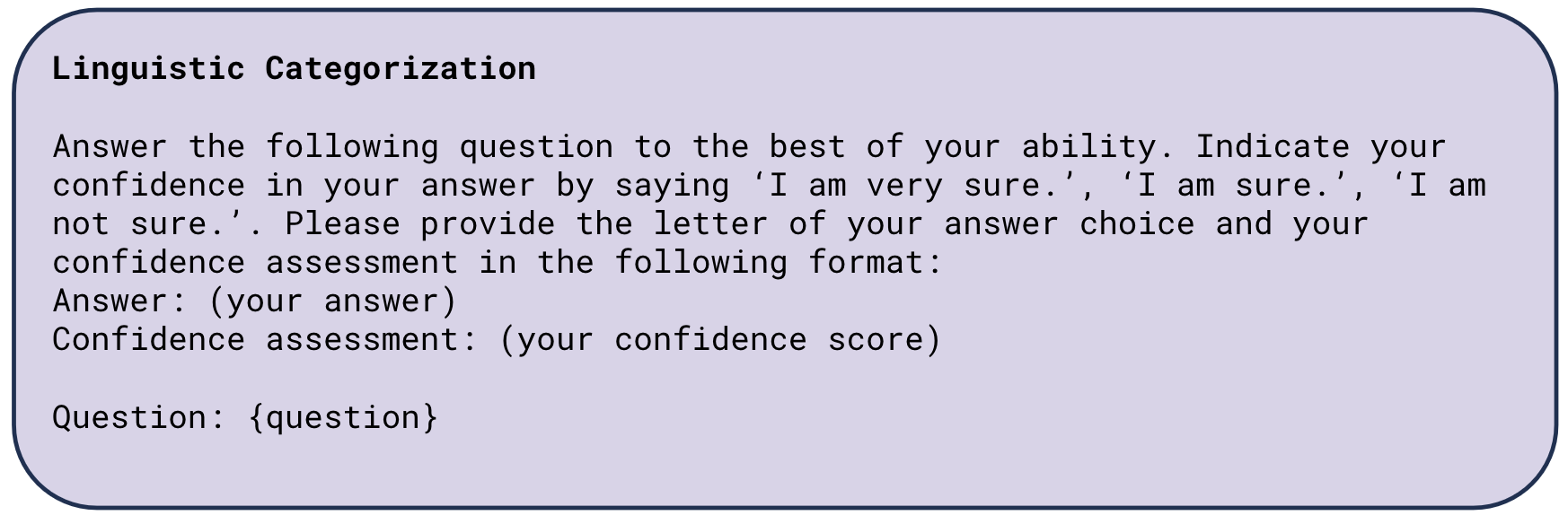}
\caption{\centering\textbf{Linguistic Categorization Prompt.} Elicits confidences in different categories of linguistic certainty.}
\end{figure}
\begin{figure}[H]
\centering
\includegraphics[scale=0.45]{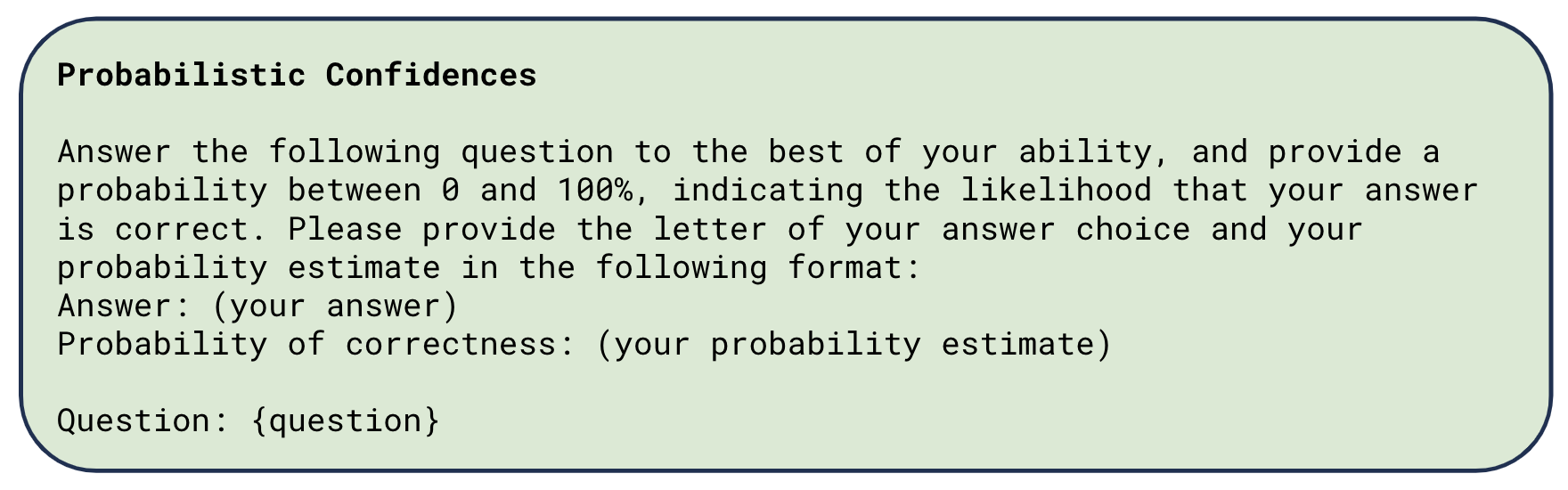}
\caption{\centering\textbf{Probabilistic Confidence Prompt.} Frames confidence as a probability between 0-100\% - in case models have seen more instances of probabilistic confidences in their training data.}
\end{figure}
\begin{figure}[H]
\centering
\includegraphics[scale=0.45]{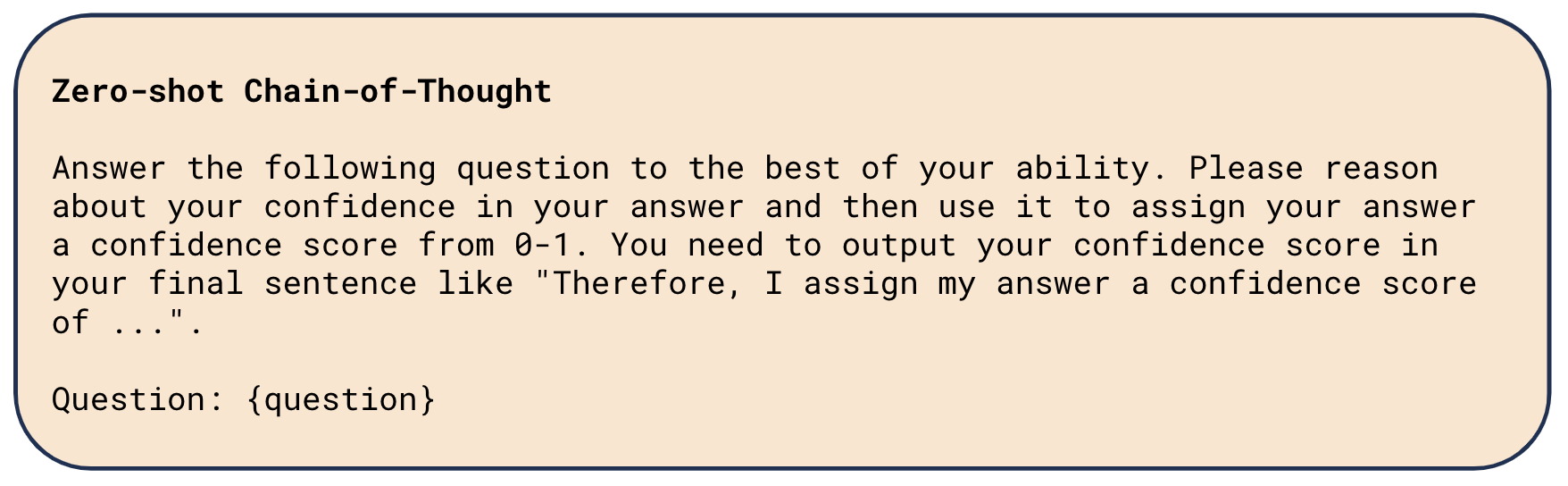}
\caption{\centering\textbf{Zero-shot Chain-of-Thought.} Allows the model to provide a chain of reasoning before deciding on its confidence score. Initial analysis indicated that these reasoning chains served to further reinforce the model's overconfidence by often incorrectly justifying its answers.}
\end{figure}

\subsection{Llama 2 70B Chat Results}
\label{llama-2-70b-results}
We also experiment with the chat version of the Llama 2 70B model, evaluating AUC and AUROC for linguistic confidences and model probabilities. We find that the chat version of the model generally performs similarly to the base version , so primary results are reported on the base model. 
Similar to Llama 2 Base, Llama 2 Chat's probabilities outperform its linguistic confidences based on AUC and AUROC on all 12 datasets. Between the chat and base versions, the base version generally outperforms chat for both linguistic confidences and model probabilities.
\begin{table}[!ht]
\begin{tabular}{@{}lllllllll@{}}
\toprule
Metric & Confidence & TQA & Medqa & CSQA & OBQA & Law & Ethics & Physics \\ \midrule
\multirow{2}{*}{AUC} & Linguistic & 0.631 & 0.521 & 0.679 & 0.750 & 0.529 & 0.675 & 0.556 \\
 & Prob & \textbf{0.699} & \textbf{0.604} & \textbf{0.840} & \textbf{0.869} & \textbf{0.674} & \textbf{0.823} & \textbf{0.721} \\ \midrule
\multirow{2}{*}{AUROC} & Linguistic & 0.683 & 0.517 & 0.506 & 0.535 & 0.501 & 0.562 & 0.568 \\
 & Prob & \textbf{0.754} & \textbf{0.609} & \textbf{0.764} & \textbf{0.776} & \textbf{0.710} & \textbf{0.821} & \textbf{0.721} \\ \bottomrule
\end{tabular}
\centering
\begin{tabular}{@{}llllllll@{}}
\toprule
Metric & Confidence & Econ & Algebra & Chem & Security & Policy & Avg \\ \midrule
\multirow{2}{*}{AUC} & Linguistic & 0.367 & 0.296 & 0.445 & 0.718 & 0.846 & 0.584 \\
 & Prob & \textbf{0.438} & \textbf{0.348} & \textbf{0.632} & \textbf{0.850} & \textbf{0.963} & \textbf{0.705} \\ \midrule
\multirow{2}{*}{AUROC} & Linguistic & 0.553 & 0.485 & 0.546 & 0.560 & 0.479 & 0.541 \\
 & Prob & \textbf{0.634} & \textbf{0.495} & \textbf{0.721} & \textbf{0.811} & \textbf{0.858} & \textbf{0.723} \\ \bottomrule
\end{tabular}
\caption{\textbf{AUC and AUROC Metrics for Llama 2 70B Chat}. Llama 2 Base's linguistic confidence scores outperform Llama 2 Chat's linguistic confidences --- 73.1\% AUC compared to 70.5\%. Similarly, Llama 2 Base's model probabilities also outperform Llama 2 Chat's probabilities --- 73.1\% AUC compared to 70.5\%. These results may support the conclusion that base models are better calibrated than chat models.}
\vspace{-0.2in}
\end{table}
\subsection{ECE Results}
\label{sec:ece_results}
Following are the ECEs for the linguistic confidence scores of each model and the ECEs of model probabilities for models which provide them.
\begin{table}[H]
\begin{tabular}{@{}llllllll@{}}
\toprule
Confidence Type & TQA & Medqa & CSQA & OBQA & Law & Ethics & Physics \\ \midrule
Text-davinci Linguistic & \textbf{0.422} & \textbf{0.425} & \textbf{0.161} & \textbf{0.127} & \textbf{0.380} & \textbf{0.300} & \textbf{0.299} \\
Text-davinci Prob & 0.461 & 0.454 & 0.235 & 0.191 & 0.388 & 0.338 & 0.338 \\ \midrule
Llama 2 Linguistic & 0.365 & 0.248 & 0.201 & \textbf{0.073} & 0.224 & 0.259 & 0.267 \\
Llama 2 Prob & \textbf{0.099} & \textbf{0.084} & \textbf{0.176} & 0.235 & \textbf{0.115} & \textbf{0.145} & \textbf{0.094} \\ \midrule
Llama 2 Chat Linguistic & 0.357 & 0.391 & 0.125 & 0.101 & 0.350 & \textbf{0.194} & 0.337 \\
Llama 2 Chat Prob & \textbf{0.284} & \textbf{0.228} & \textbf{0.124} & \textbf{0.092} & \textbf{0.264} & 0.213 & \textbf{0.210} \\ \midrule
GPT-3.5 Linguistic & 0.350 & 0.380 & 0.192 & 0.091 & 0.388 & 0.176 & 0.363 \\
Claude-v1.3 Linguistic & 0.187 & \textbf{0.086} & \textbf{0.042} & \textbf{0.033} & \textbf{0.098} & \textbf{0.052} & 0.162 \\
GPT-4 Linguistic & \textbf{0.104} & 0.118 & 0.118 & 0.038 & 0.187 & 0.114 & \textbf{0.109} \\ \bottomrule
\end{tabular}
\centering
\begin{tabular}{@{}lllllll@{}}
\toprule
Confidence Type & Econ & Algebra & Chem & Security & Policy & Avg \\ \midrule
Text-davinci Linguistic & 0.482 & 0.625 & 0.475 & \textbf{0.213} & \textbf{0.038} & \textbf{0.329} \\
Text-davinci Prob & \textbf{0.478} & \textbf{0.576} & \textbf{0.385} & 0.263 & 0.112 & 0.352 \\ \midrule
Llama 2 Linguistic & 0.453 & 0.561 & 0.435 & \textbf{0.079} & \textbf{0.093} & 0.272 \\
Llama 2 Prob & \textbf{0.205} & \textbf{0.091} & \textbf{0.100} & 0.172 & 0.264 & \textbf{0.148} \\ \midrule
Llama 2 Chat Linguistic & 0.505 & 0.480 & 0.480 & \textbf{0.165} & \textbf{0.055} & 0.295 \\
Llama 2 Chat Prob & \textbf{0.403} & \textbf{0.361} & \textbf{0.272} & 0.187 & 0.073 & \textbf{0.226} \\ \midrule
GPT-3.5 Linguistic & 0.515 & 0.560 & 0.432 & 0.173 & \textbf{0.094} & 0.309 \\
Claude-v1.3 Linguistic & \textbf{0.132} & \textbf{0.319} & \textbf{0.175} & \textbf{0.058} & 0.162 & \textbf{0.126} \\
GPT-4 Linguistic & 0.270 & 0.420 & 0.313 & 0.118 & 0.053 & 0.164 \\ \bottomrule
\end{tabular}
\caption{\textbf{ECE Values Linguistic Confidences vs Model Probabilities} The ECE values for linguistic confidences and model probabilities show that on some datasets model probabilities achieve better ECE values, while on other datasets linguistic confidences achieve better ECE values. Among the state-of-the-art models, Claude-v1.3's linguistic confidences notably result in the least expected calibration error on 9/12 datasets.}
\end{table}
\textbf{Metric.} 
We compute the \textit{expected calibration error metric (ECE)} by dynamically binning examples based on their confidence scores into 10 bins with approximately equal numbers of examples in each bin. For each bin, we compute the calibration error, which is the absolute difference between the mean predicted confidence and the mean observed accuracy. This quantifies how well the predicted confidences match the true probability of correctness within each bin. We then calculate the weighted average of the calibration errors across all bins, where the weights are the proportion of examples in each bin relative to the total number of examples. 

We also compute the ECE values for our baseline confidence methods (linguistic confidences, SC linguistic confidences) and for our proposed confidence methods (surrogate, tiebreak, mixture, and SC mixture) for the GPT-4 model. For 11 out of 12 tasks, we find that our proposed methods lead to the lowest ECE values.
\begin{table}[H]
\vspace{-0.1in}
\begin{tabular}{@{}llllllll@{}}
\toprule
 & TQA & MedQA & CSQA & OBQA & Law & Ethics & Physics \\ \midrule
 \rowcolor{grayrowcolor}
Ling. Conf. & 0.104 & 0.118 & 0.118 & 0.038 & 0.187 & 0.114 & 0.109 \\
\rowcolor{grayrowcolor}
SC Ling. Conf. & 0.126 & 0.163 & 0.120 & 0.036 & 0.246 & 0.204 & 0.120 \\
Surrogate$^\dagger$ & 0.395 & 0.212 & 0.297 & 0.370 & 0.156 & 0.205 & 0.317 \\
Tiebreak$^\dagger$ & 0.114 & 0.134 & 0.126 & \textbf{0.032} & 0.194 & 0.114 & 0.118 \\
Mixture$^\dagger$ & 0.096 & \textbf{0.075} & \textbf{0.061} & 0.159 & \textbf{0.064} & \textbf{0.111} & \textbf{0.088} \\
SC Mixture$^\dagger$ & \textbf{0.085} & 0.120 & 0.108 & 0.029 & 0.216 & 0.186 & 0.098 \\ \bottomrule
\end{tabular}
\centering
\begin{tabular}{@{}lllllll@{}}
\toprule
Confidence Type & Econ & Algebra & Chem & Security & Policy & Avg \\ \midrule
\rowcolor{grayrowcolor}
Ling. Conf. & 0.270 & 0.420 & 0.313 & 0.118 & \textbf{0.053} & 0.164 \\
\rowcolor{grayrowcolor}
SC Ling. Conf. & 0.323 & 0.379 & 0.331 & 0.136 & 0.063 & 0.187 \\
Surrogate$^\dagger$ & 0.129 & \textbf{0.162} & \textbf{0.187} & 0.210 & 0.264 & 0.242 \\
Tiebreak$^\dagger$ & 0.270 & 0.419 & 0.332 & 0.158 & 0.068 & 0.173 \\
Mixture$^\dagger$ & \textbf{0.126} & 0.224 & 0.229 & \textbf{0.108} & 0.138 & \textbf{0.123} \\
SC Mixture$^\dagger$ & 0.287 & 0.358 & 0.286 & 0.129 & 0.068 & 0.164 \\ \bottomrule
\end{tabular}
\caption{\textbf{ECE Values All Confidence Methods for GPT-4}}
\end{table}
\subsection{Surrogate Model Results} 
\label{sec:surrogate_model_details}
\begin{figure}[H]
\centering
    \includegraphics[scale=0.52]{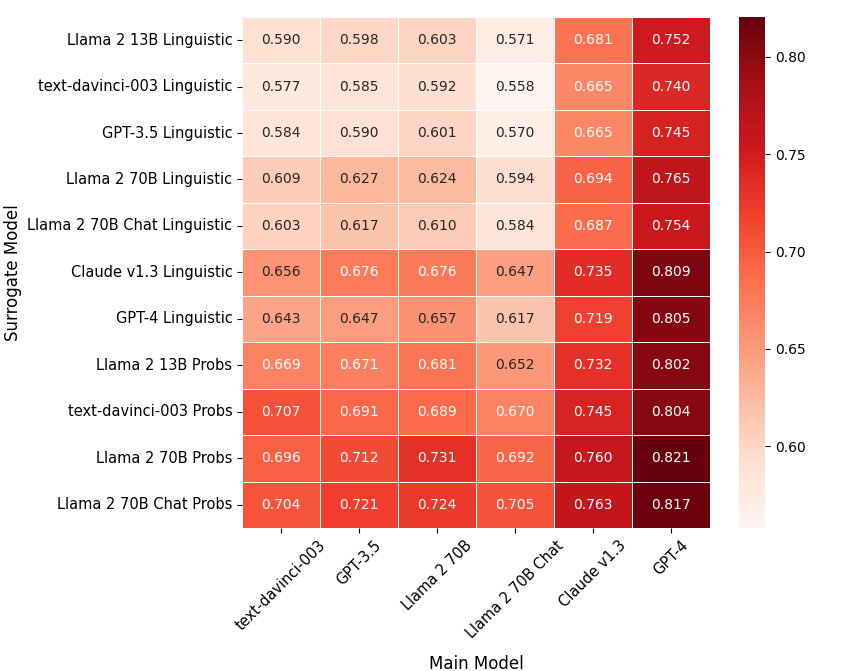}
    \caption{\textbf{AUCs For All Surrogate Models} We compute the AUC metric for each model considering surrogate confidences from both model probabilities and linguistic confidence scores from all other models. We find that all models benefit from using surrogate model probabilities over their own linguistic confidences.}\label{fig:full_surrogate_auc_heatmap}
\end{figure}
\begin{figure}[H]
\centering
    \includegraphics[scale=0.52]{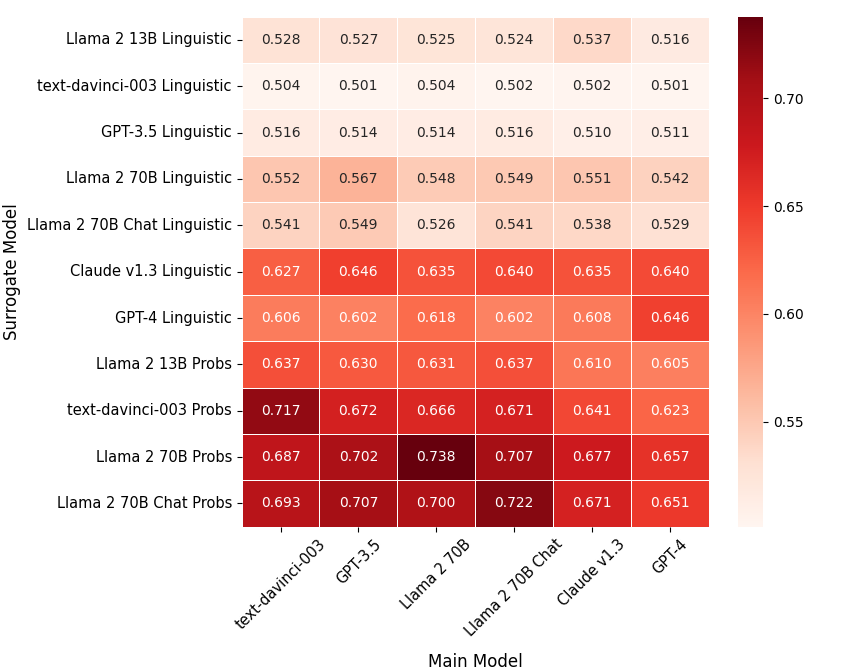}
    \caption{\textbf{AUROCs For All Surrogate Models} We also compute the AUROC metric for each model considering surrogate confidences from both model probabilities and linguistic confidence scores from all other models. In general, we find that using surrogate model probabilities instead of a model's own linguistic confidences improves AUROC values.}
    \label{fig:fullsurrogate_auroc_heatmap}
\end{figure}
AUCs and AUROCs for surrogate models show that model probabilities from other models can provide better confidence estimates than a models own linguistic confidences.
\subsection{Correlation and Covariance of Surrogate and Main Models}
\label{sec:analysis_appendix}
\begin{figure}[H]
\centering
    \includegraphics[scale=0.52]{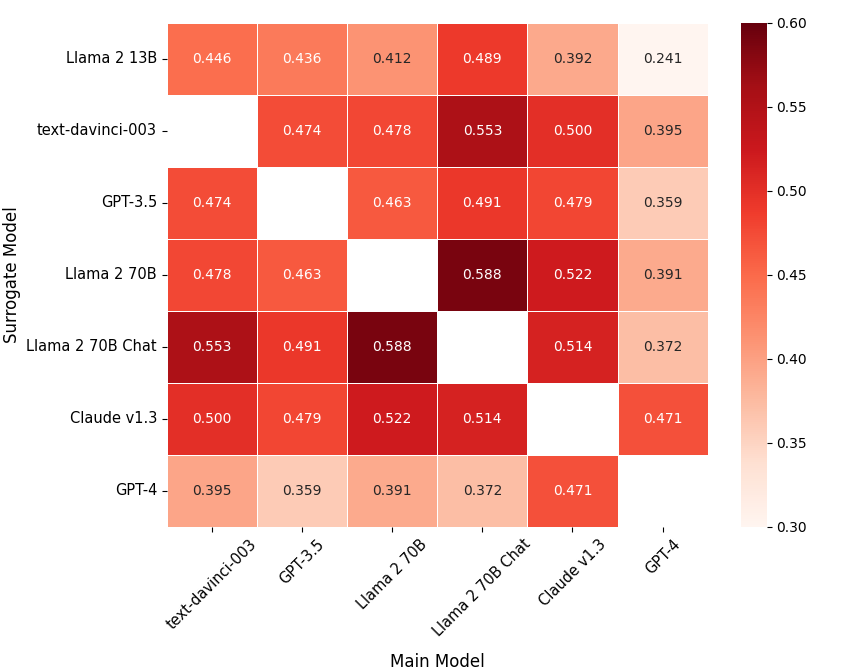}
    \caption{\textbf{Correlations For All Main and Surrogate Models}}
    \label{fig:corr_surrogate_heatmap}
\end{figure}
\begin{figure}[H]
\centering
    \includegraphics[scale=0.52]{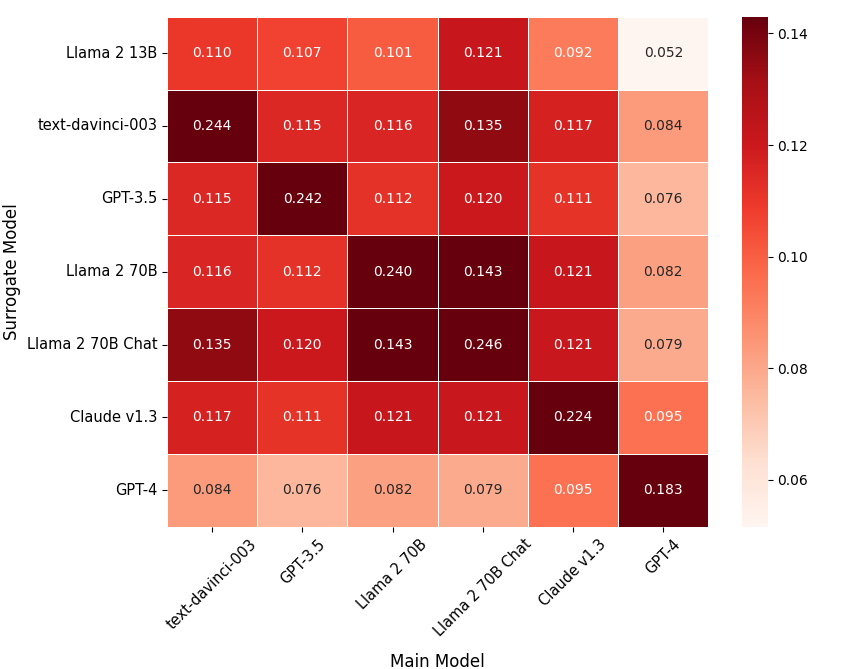}
    \caption{\textbf{Covariances For All Main and Surrogate Models}}
    \label{fig:cov_surrogate_heatmap}
\end{figure}
\vspace{-0.2in}
We compute correlations and covariances between the answer correctness (set of binary scores indicating if a model answered a question correctly or not) for every pair of main model and potential surrogate model. We find that in general if a surrogate model $S$ has a high degree of correlation in answer correctness with a main model $M$, then $S$ is likely to be a good surrogate for $M$. For example, GPT-4 has a higher correlation with Llama 2 Base, Llama 2 Chat, and text-davinci-003 than it does with Llama 2 13B indicating that those models can be better surrogates for GPT-4 than Llama 2 13B.  Similar trends also hold for covariances.
\subsection{Calibration of Mixture of Models}
\begin{figure}[H]
    \centering
    \begin{subfigure}{0.32\linewidth}
        \includegraphics[width=\linewidth]{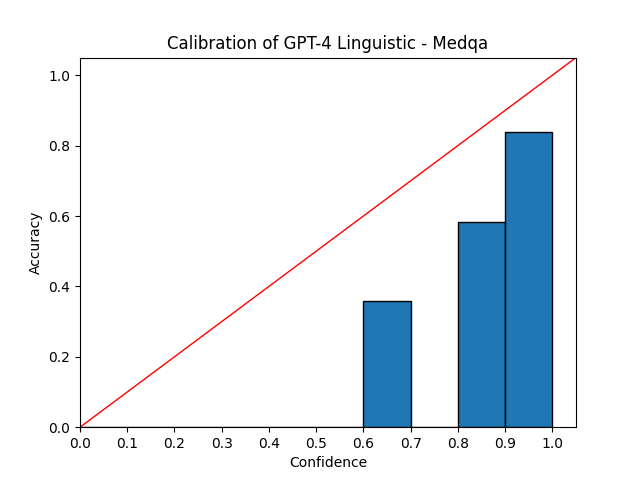}
    \end{subfigure}
    \begin{subfigure}{0.32\linewidth}
        \includegraphics[width=\linewidth]{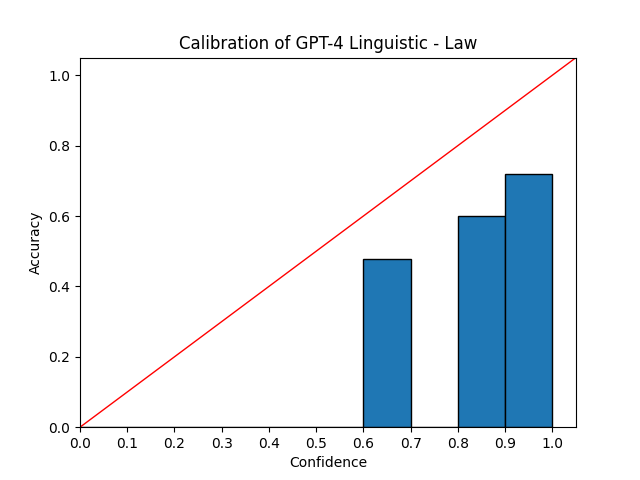}
    \end{subfigure}
    \begin{subfigure}{0.32\linewidth}
        \includegraphics[width=\linewidth]{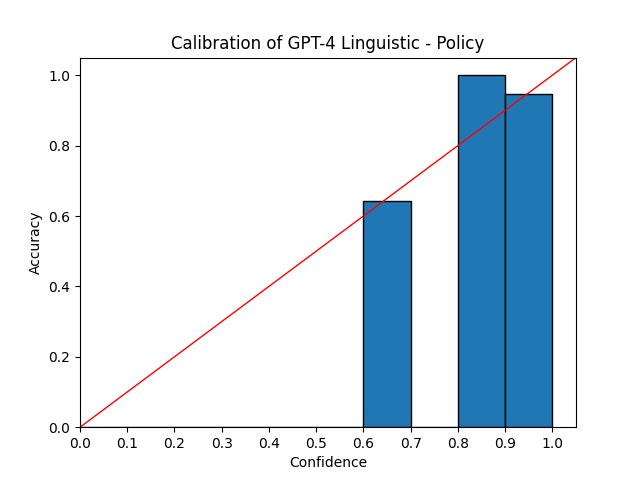}
    \end{subfigure}
    \begin{subfigure}{0.32\linewidth}
        \includegraphics[width=\linewidth]{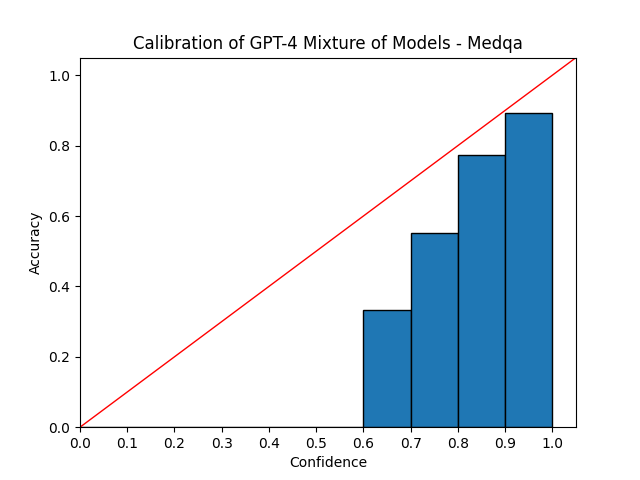}
    \end{subfigure}
    \begin{subfigure}{0.32\linewidth}
        \includegraphics[width=\linewidth]{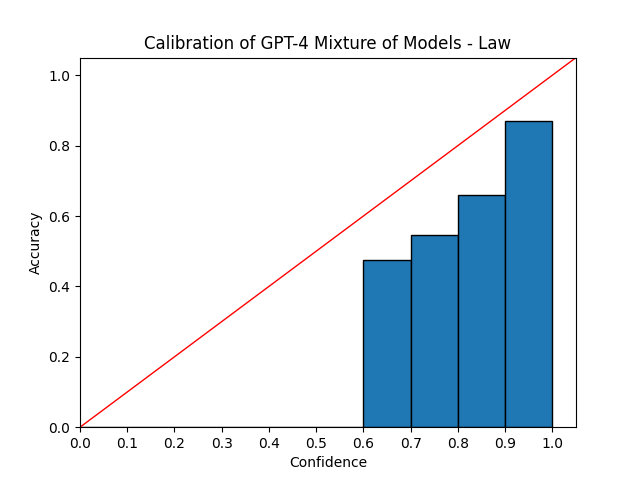}
    \end{subfigure}
    \begin{subfigure}{0.32\linewidth}
        \includegraphics[width=\linewidth]{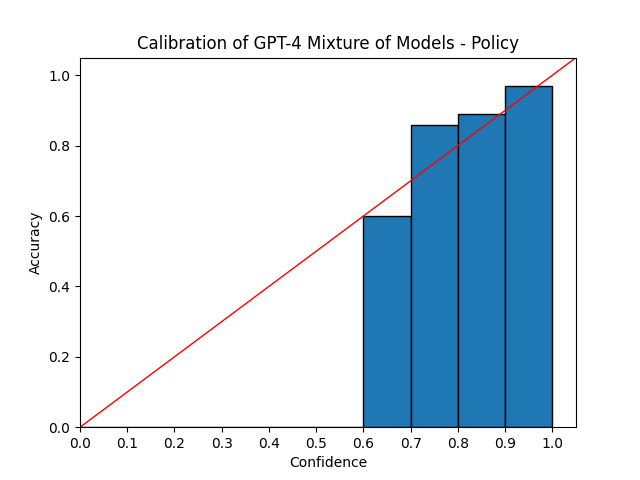}
    \end{subfigure}   
    \caption{\textbf{Calibration of GPT-4 with Linguistic Confidence and Mixture of Models} In the first row we see the calibration of GPT-4 on MedQA, Professional Law, and US Foreign Policy when using linguistic confidences. In the second row, we see GPT-4's calibration using our Mixture of Models confidence method. A perfectly calibrated model would have all bars aligned with the red line (average confidence in each bucket is exactly equal to the average accuracy). We can see that calibration improves demonstrably, when using Mixture of Models.}
\end{figure}